\documentclass{ieeeaccess}
\usepackage{cite}
\usepackage{amsmath,amssymb,amsfonts}
\usepackage{algorithmic}
\usepackage{graphicx}
\usepackage{textcomp}
\usepackage{times}
\usepackage{epsfig}

\usepackage[utf8]{inputenc} 
\usepackage[T1]{fontenc}    
\usepackage{url}            
\usepackage{booktabs}       
\usepackage{amsfonts}       
\usepackage{nicefrac}       
\usepackage{microtype}      
\usepackage{algorithm}
\usepackage{multirow}
\usepackage{mathtools}
\usepackage{color}
\usepackage{subfigure}

\usepackage[caption=false]{subfig}
\usepackage{caption,setspace}
\usepackage{textcomp}
\def\BibTeX{{\rm B\kern-.05em{\sc i\kern-.025em b}\kern-.08em
    T\kern-.1667em\lower.7ex\hbox{E}\kern-.125emX}}

\begin{document}
\history{Date of publication December 00, 2020, date of current version December 1, 2020.}
\doi{10.1109/ACCESS.2020.3042302}

\title{Joint Learning of Generative Translator and Classifier for Visually Similar Classes}
\author{\uppercase{ByungIn Yoo}\authorrefmark{1,2},
\uppercase{Tristan Sylvain}\authorrefmark{3},
\uppercase{Yoshua Bengio}\authorrefmark{4}, and \uppercase{Junmo Kim}\authorrefmark{5}
}
\address[1]{School of Electrical Engineering, Korea Advanced Institute of Science and Technology (e-mail: byungin.yoo@kaist.ac.kr)}
\address[2]{Computer Vision Lab., Samsung Advanced Institute of Technology (e-mail: byungin.yoo@samsung.com)}
\address[3]{Montreal Institute for Learning Algorithms (e-mail: tristan.sylvain@gmail.com)}
\address[4]{Montreal Institute for Learning Algorithms (e-mail: yoshua.bengio@mila.quebec)}
\address[5]{School of Electrical Engineering, Korea Advanced Institute of Science and Technology (e-mail: junmo.kim@kaist.ac.kr)}


\markboth
{Author \headeretal: Preparation of Papers for IEEE TRANSACTIONS and JOURNALS}
{Author \headeretal: Preparation of Papers for IEEE TRANSACTIONS and JOURNALS}

\corresp{Corresponding author: junmo.kim (e-mail: junmo.kim@kaist.ac.kr).}

\begin{abstract}
In this paper, we propose a Generative Translation Classification Network (GTCN) for improving visual classification accuracy in settings where classes are visually similar and data is scarce.
For this purpose, we propose joint learning from a scratch to train a classifier and a generative stochastic translation network end-to-end.
The translation network is used to perform on-line data augmentation across classes, whereas previous works have mostly involved domain adaptation.
To help the model further benefit from this data-augmentation, we introduce an adaptive fade-in loss and a quadruplet loss.
We perform experiments on multiple datasets to demonstrate the proposed method's performance in varied settings.
Of particular interest, training on 40\% of the dataset is enough for our model to surpass the performance of baselines trained on the full dataset.
When our architecture is trained on the full dataset, we achieve comparable performance with state-of-the-art methods despite using a light-weight architecture.
\end{abstract}


\titlepgskip=-15pt
\maketitle

\section{Introduction}
\label{sec:introduction}
\PARstart{G}{enerative} models have received significant interest in the past years.
Although recent models can generate realistic and diverse data,
more study is needed to ascertain whether such methods can also be useful in enhancing classification accuracy on hard settings such as when classes are visually similar, or there is a scarcity of training data.
For example, face liveness detection in biometrics is a crucial problem where it is hard to distinguish between real faces and printed fake faces,
because examples from the two classes are very similar \cite{akbulut2017deep, atoum2017face, tian2016spoofing, tang2018face, song2018face, li2018learning, liu2018learning}.
In this application, obtaining a high true acceptance ratio (TAR) and a low false acceptance ratio (FAR) is important as a high TAR is essential for user convenience, whereas a low FAR results in better security.
This paper is motivated by two questions:
\begin{itemize}
\item If two classes, $A$ and $B$, are visually very similar, how can we improve classifiers by employing cross-class generative models?
\item When data is scarce, how can we use generative models to learn better representations for the visually similar classes?
\end{itemize}

In practice, most past approaches to solving these two questions fall in two categories. The first one is using complex models, which are often hard to train, and make it difficult to perform fast inference in settings with limited computing resources, such as on a smartphone. The second is to collect large amounts of training data, which is costly, time-consuming, and not always straightforward.
In this paper, we propose a Generative Translation Classification Network that uses the translation model of visual classes to assist the training of the classifier via exploiting joint learning.
If the translation model is able to effectively augment the quantity of training samples we expect both the issue of having closely distributed classes, and the lack of sufficient amounts of training data to be mitigated.
We should note the similarity of our proposed method with the way the brain learns.
There is biological evidence that long-term memory is formed by the collaboration between the hippocampus and the prefrontal cortex \cite{preston2013interplay}. The hippocampus recalls slightly distorted memories, in a way which inspired our translation network generating variations on training examples. The prefrontal cortex uses such memories, and has a role analogous to the classifier we used.
In summary, our contributions are the following:
\begin{itemize}
\item To augment mini-batch data (AMB) during training, we use inter-class translated samples based on joint learning of a translation model and a classifier.
    Specifically, half the training samples seen by the classifier are inter-class translated samples that are stochastically generated (ST), while the other half of the samples are real samples of training data to preserve the original data distribution.
    This is a novel attempt to couple a generative translation network and a classifier in a unified architecture for improving classification of visually similar classes.
\item Early on during joint learning, translated samples are of poor quality. We use adaptive fade-in training (AF) for the classifier, automatically adjusting the importance of real and translated samples to gradually adapt the influence from translated samples.
    We design a novel quadruplet loss (QL) that helps preserve intra-class distribution and taking inter-class distribution apart, even though generated samples are used for training.
    This is due to the fact that this loss encourages similarity between the embeddings of real and intra-class translated images, and dissimilarity between the embeddings of inter-class samples.
\end{itemize}

\begin{figure}[!t] 
  \centering
  \centerline{\includegraphics[width=7cm]{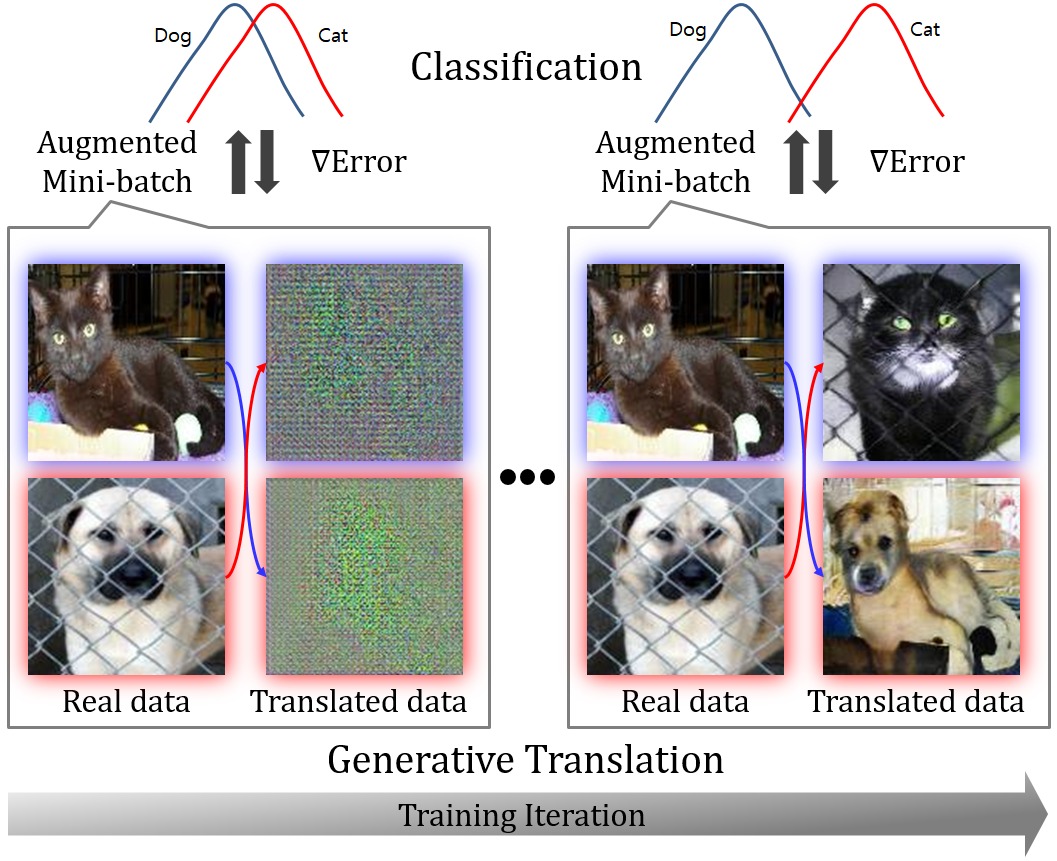}}
  \caption{A generative translation model progressively generates samples to augment data in  mini-batch for joint training of a classifier.}
  \label{fig:Concept}
\end{figure}

\section{Related work}\label{sec:relatedwork}
Since the introduction of generative adversarial networks (GANs) \cite{goodfellow2014generative},
many GAN-based generative models have been introduced, e.g. \cite{mirza2014conditional, chen2016infogan, odena2016conditional, antoniou2017data, creswell2018generative}.
A typical issue is mode collapse, reduced diversity of generated samples, which would be detrimental to improve classification accuracy in our setup.
In addition to this, our specific need for inter-class generated samples is difficult to meet with classical GANs, hence we focus on adversarial translation models.
Domain translation has received renewed interest in recent years, thanks in large part to the use of adversarial methods.
The first results using adversarial training used paired samples~\cite{isola2017image}. More recent developments using cycle consistency allow for unsupervised domain translation using unpaired samples~\cite{taigman2016unsupervised, zhu2017unpaired, liu2017unsupervised}.
Finally, newer methods extend this approach to problems requiring many-to-many mappings or multi-modal data distributions~\cite{choi2017stargan, zhu2017toward, huang2018multimodal, almahairi2018augmented, DRIT}.
The power of these approaches lies in being able to learn such transformations without requiring examples of a one-to-one mapping between training data in source and target domains.

Multiple works have used the discriminators of GANs as semi-supervised learning classifiers \cite{springenberg2015unsupervised, dumoulin2016adversarially, gan2017triangle, chongxuan2017triple, dai2017good}.
Typically, the semi-supervised classifiers take a tiny portion of labeled data and a much larger amount of unlabeled data from the same domain. The goal is to use both labeled and unlabeled data to train a neural network so that it can map a new data point to its correct class \cite{odena2016semi}.
A simple yet effective idea for semi-supervised learning is to turn a classification problem with $n$ classes into a classification problem with $n+1$ classes, with the additional class corresponding to fake images \cite{salimans2016improved}.
However, the methods are not directly augmenting data of existing classes, since they assume generated data is a new class.

The feature matching loss \cite{salimans2016improved} was introduced as an additional loss term to prevent instability of GANs from over-fitting on the discriminator.
Specifically, the generator is trained to match the expected value of the features on an intermediate layer of the discriminator.
While feature matching losses bring benefits, a cost function that can account for inter- and intra-class relationships is required in our case.
Deep learning networks with variants of a triplet loss become common methods for face verification \cite{schroff2015facenet} and person re-identification \cite{chen2017beyond}.
Despite this interest, such losses have seldom been applied to generative models.

From another perspective, between-class data augmentation methods have been proposed \cite{tokozume2017between}.
Images are generated by mixing two training images belonging to different classes with a random ratio.
The training procedure seeks to minimize the KL-divergence between the outputs of a trained model, and a target computed by interpolating the two one-hot target vectors of the initial examples using the same ratio.
Even though the methods achieve superior results on visual recognition,
it is not clear whether the classifier learns shaper and more diverse data on visually similar classes.

Data subrogation is an important technique that a reduced set of real credit card data is used to generate surrogate multivariate series of credit card data transactions~\cite{salazar2014surrogate}.
The surrogate multivariate data are constrained to have the same covariance, marginal distributions, and joint distributions as the original multivariate data.
The surrogate data are combined with real data to consist of the mixed dataset.
The detection results obtained from the mixed dataset are comparable to those obtained from only real data,
while the detection accuracy of the mixed dataset doesn’t outperform that of the real dataset.

In this paper, we use translated images obtained via a generative model to perform data augmentation in each mini-batch and train a classifier.
Even though many previous works have applied generative models to classifier training~\cite{sylvain2019locality},
this work is first to focus on joint learning to improve classification accuracy for visually similar images.

\begin{figure*}[!ht]
  \centering
  \centerline{
    {\includegraphics[width=11cm]{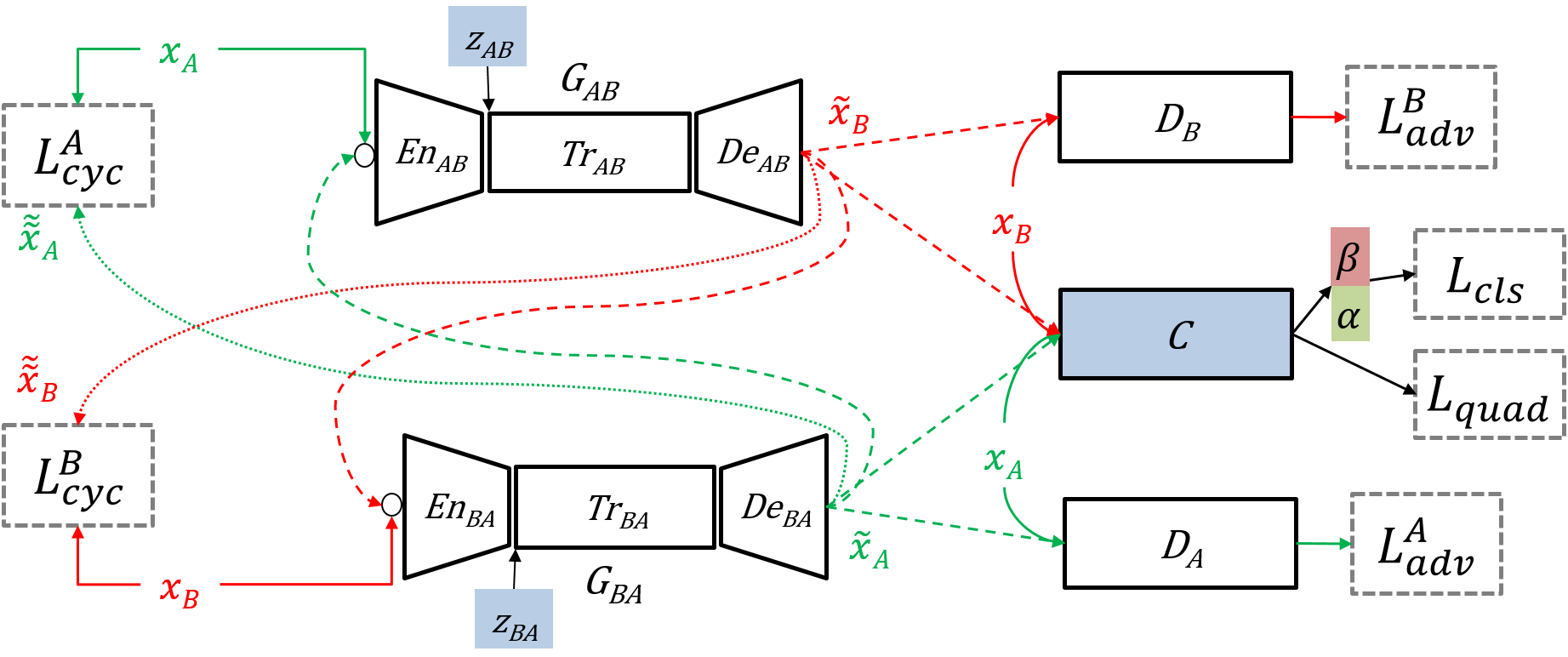}}
    \hspace{10mm}
    {\includegraphics[width=4.5cm]{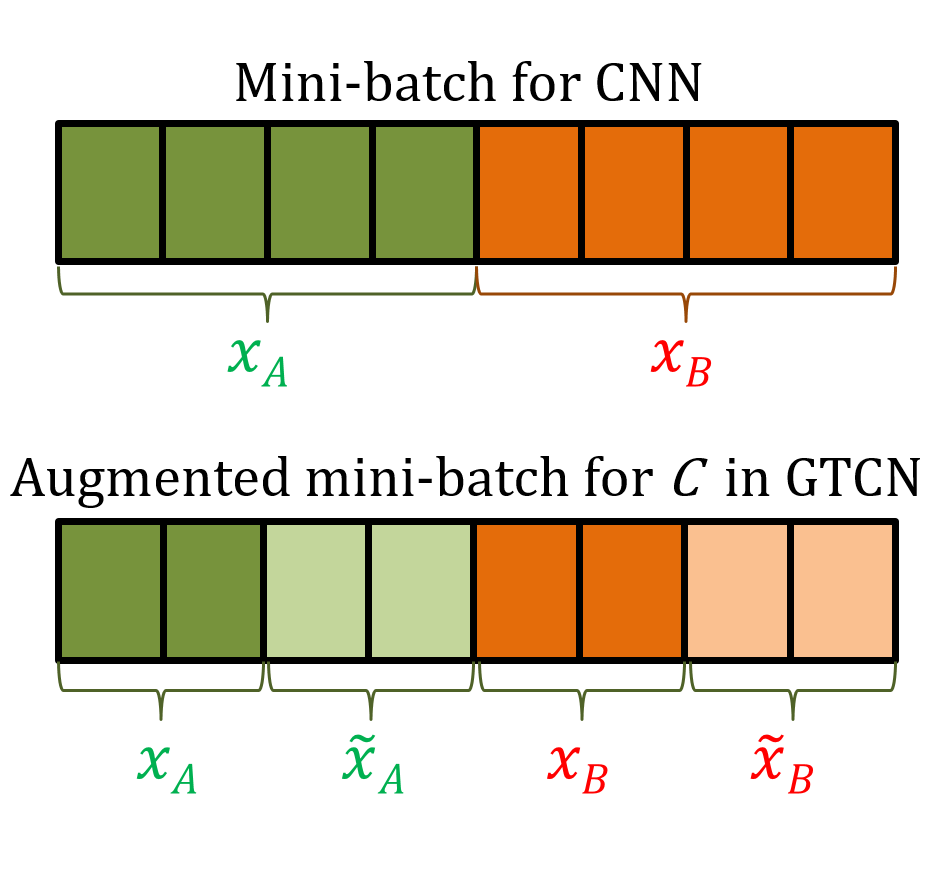}}
  }
  \caption{Overview of the proposed Generative Translation Classification Network :
  (Left) $G_{AB}$ is a translator from class $A$ to $B$, while $G_{BA}$ is a corresponding translator from class $B$ to $A$.
  $G_{AB}$ consists of three main blocks, an encoder $En_{AB}$ (three convolutional layers), transformer $Tr_{AB}$ (nine residual blocks), and decoder $De_{AB}$ (two deconvolutional layers and one convolutional layer).
  Likewise, $G_{BA}$ consists of $En_{BA}$, $Tr_{BA}$, and $De_{BA}$.
  Random noise $\{z_{AB}, z_{BA}\}\in\mathbb{R}^{32\times32\times3}$ for stochastic translation is sampled from a uniform distribution over $[-1,1]$ and concatenated to the output maps of $En_{AB}$ and $En_{BA}$.
  $D_{A}$ and $D_{B}$ are the corresponding discriminators of adversarial training for classes A and B respectively.
  $C$ is a simple classifier consisting of six convolutional layers over classes $A$ and $B$.
  $\alpha$ and $\beta$ are the trainable parameters of the adaptive fade-in loss.
  $x_A$ and $x_B$ are real samples from classes A and B respectively.
  The corresponding translated samples are denoted by $\tilde{x}_{B}$ and $\tilde{x}_{A}$.
  The cyclic reconstructed samples are $\tilde{\tilde{x}}_{A}=G_{BA}(\tilde{x}_{B})$ and $\tilde{\tilde{x}}_{B}=G_{AB}(\tilde{x}_{A})$.
  Note that only $C$ is used at test-time.
  Details of loss functions $L{*}$ are in proposed methods section. 
  (Right) Comparison of mini-batch structure. A mini-batch used for training $C$ in GTCN contains real and translated samples, while the baseline CNNs are only trained on real samples.
  }
  \label{fig:overview}
\end{figure*}

\section{Proposed methods}\label{sec:approaches}
We design a unified deep network architecture that combines a classifier $C$ and a generative translation model $G$ to perform on-line data augmentation of the mini-batch.
The proposed architecture is explained in Figure ~\ref{fig:overview}.

\subsection{Formulation of proposed methods}
Let $\{ x_{y}^{i} : 1 \leq i \leq n, y \in Y \}$ be a dataset such that $x_{y}^{i}$$\in$$X$ is the $i$-th sample belonging to class $y\in Y$.
We consider a learning algorithm that trains a classifier
\small
\begin{align}\label{eq:joint_define}
C:X\rightarrow Y
\end{align}
\normalsize
by jointly using a generative translation model $G(\tilde{x}|x)$.
Our goal is to improve the classifier $C$ by utilizing the on-line translated samples $\tilde{x}$ as additional training data.

More formally, $C$ is a classifier to discriminate two-class cases, where $Y$=$\{A,B\}$.
Generative translation models $G$ are employed to produce $\tilde{x}$ from $x$ as:
\small
\begin{align}\label{eq:translate_define}
\tilde{x}_{A}^{i}=G_{BA}({x}_{B}^{i}, z_{BA})~,\\
\tilde{x}_{B}^{i}=G_{AB}({x}_{A}^{i}, z_{AB})~,
\end{align}
\normalsize
where $x_{A}^{i}$ and $x_{B}^{i}$ are real samples of class $A$ and $B$,
$z_{BA}$ and $z_{AB}$ are tensors of random noise used for stochastic translation,
and $\tilde{x}_{A}^{i}$ and $\tilde{x}_{B}^{i}$ are translated samples across the classes.
One of our important contributions is to add a translator into the full pipeline for a classifier and to jointly train the translator and the classifier.
Namely, $G_{AB}$ and $G_{BA}$ provide diverse and challenging data continuously for training classifier $C$.
So, we suppose $G_{AB}$ and $G_{BA}$ generate effective images from the standpoint of decision boundary of the classifier.

\subsubsection{Definition of augmented mini-batch}
The augmented mini-batch that is input to the classifier $C$ during training is defined as follows:
\small
\begin{align}\label{eq:amb_define}
AMB_{k}=\{x_{A}^{i}\}_{i=1}^{m}\cup\{\tilde{x}_{A}^{i}\}_{i=1}^{m}\cup\{x_{B}^{i}\}_{i=1}^{m}\cup\{\tilde{x}_{B}^{i}\}_{i=1}^{m}~,
\end{align}
\normalsize
where $k$ is the training iteration, $m$ is the number of samples in each set.
The $\tilde{x}$'s in $AMB_{k}$ increase diversity of training data, while $x$'s preserve original distribution of training data.

\subsubsection{Loss of stochastic translator}
Visual translation objective across classes is that a source class borrow underlying structure from a target class, while maintain style of the source class.
To meet the needs, cycle consistency losses, $L_{cyc}^{A}$ and $L_{cyc}^{B}$, are used to train stochastic translators, $G_{AB}$ and $G_{BA}$.
The objective is expressed as:
\small
\begin{align}\label{eq:loss_cyc_a}
L_{cyc}^{A}=\mathbb{E}_{x_{A}\sim P_{data}}[\|G_{BA}(\tilde{x}_{B},z_{BA})-x_{A}\|_{1}]~,
\end{align}
\normalsize
\small
\begin{align}\label{eq:loss_cyc_b}
L_{cyc}^{B}=\mathbb{E}_{x_{B}\sim P_{data}}[\|G_{AB}(\tilde{x}_{A},z_{AB})-x_{B}\|_{1}]~.
\end{align}
\normalsize

For generating realistic images, an adversarial loss $L_{adv}^{A}$ is applied to train $D_{A}$ and $G_{BA}$.
Similarly, an adversarial loss $L_{adv}^{B}$ is applied to train $D_{B}$ and $G_{AB}$.
The objective is expressed as:
\small
\begin{align}\label{eq:loss_adv_a}
L_{adv}^{A}&=\mathbb{E}_{x_{A}\sim P_{data}}[\log(D_{A}(x_{A}))]~\notag \\
           &+\mathbb{E}_{x_{B}\sim P_{data}}[\log(1-D_{A}(G_{BA}(x_{B}, z_{BA})))]~,
\end{align}
\normalsize
\small
\begin{align}\label{eq:loss_adv_b}
L_{adv}^{B}&=\mathbb{E}_{x_{B}\sim P_{data}}[\log(D_{B}(x_{B}))]~\notag \\
           &+\mathbb{E}_{x_{A}\sim P_{data}}[\log(1-D_{B}(G_{AB}(x_{A}, z_{AB})))]~.
\end{align}
\normalsize

\subsubsection{Adaptive fade-in learning}
In the early stages, translated samples in $AMB_{k}$ are relatively poor and relying on them too much could be detrimental for training classifier $C$.
We therefore design an adaptive fade-in loss (AF) that adapts the importance given to real and translated samples during training.
The objective is defined as:
\small
\begin{align}\label{eq:loss_dcl}
L&_{cls} = \alpha\cdot\mathbb{E}_{x_{A}\sim P_{data}}[-\log(P(y=A|x_{A}))] \notag \\
 & + (1-\alpha)\cdot\mathbb{E}_{x_{B}\sim P_{data}}[-\log(P(y=A|G_{BA}(x_{B}, z_{BA})))] \notag \\
 & + \beta\cdot\mathbb{E}_{x_{B}\sim P_{data}}[-\log(P(y=B|x_{B}))] \notag \\
 & + (1-\beta)\cdot\mathbb{E}_{x_{A}\sim P_{data}}[-\log(P(y=B|G_{AB}(x_{A}, z_{AB})))]~,
\end{align}
\normalsize
where $\alpha$ and $\beta$ are parameters between 0 to 1.
We use a categorical cross-entropy loss $L_{cls}$ on the Softmax output of $C$.
The classification loss $L_{cls}$ is used to train $C$, $G_{AB}$, and $G_{BA}$ by backpropagation.
In the equation ~\ref{eq:loss_dcl}, the parameters $\alpha$ and $\beta$ control the relative importance of four training inputs,
$x_{A}$, $\tilde{x}_{A}$=$G_{BA}(x_{B}, z_{BA})$, $x_{B}$, and $\tilde{x}_{B}$=$G_{AB}(x_{A}, z_{AB})$ in $AMB_{k}$.
Specifically, $\alpha$ controls the relative weight given to real data $x_{A}$ and augmented data $\tilde{x}_{A}$ to train a classifier $C$.
Similarly, $\beta$ controls the relative importance given to real data $x_{B}$ and augmented data $\tilde{x}_{B}$ to train the classifier $C$.

\subsubsection{Quadruplet loss for inter/intra-class}
We design a quadruplet loss for the proposed architecture that enforces explicit relationships between classes.
The objective is defined as:
\small
\begin{align}\label{eq:loss_quad}
L&_{quad} = \notag \\
 & [\|f_{C}(x_{A})-f_{C}(\tilde{x}_{A})\|_{2}^{2} - \|f_{C}(x_{A})-f_{C}(\tilde{x}_{B})\|_{2}^{2} + \eta_{a}]_{+} \notag \\
 & + [\|f_{C}(x_{B})-f_{C}(\tilde{x}_{B})\|_{2}^{2} - \|f_{C}(x_{B})-f_{C}(\tilde{x}_{A})\|_{2}^{2} + \eta_{b}]_{+} \notag \\
 & + [ - \|f_{C}(x_{A})-f_{C}(x_{B})\|_{2}^{2} - \|f_{C}(\tilde{x}_{A})-f_{C}(\tilde{x}_{B})\|_{2}^{2} + \eta_{c}]_{+}~,
\end{align}
\normalsize
where $[\cdot]_{+} = max(\cdot,0)$, and $f_{C}(\cdot)$ denotes the embedding features of $x_{A}$, $\tilde{x}_{A}$, $x_{B}$, and $\tilde{x}_{B}$ in $AMB_{k}$ at the last feature layer of the classifier C.
The thresholds $\eta_{a}$, $\eta_{b}$, and $\eta_{c}$ are margins that are enforced between positive and negative pairs.
Likewise $L_{cls}$, $L_{quad}$ is used to train $C$, $G_{AB}$, and $G_{BA}$.
The quadruplet loss encourages the similarity of intra-class samples and the dissimilarity of inter-class samples, which is useful for classification.
Specifically, the intention of the designing the quadruplet loss considers all six pairs out of the four elements, $x_{A}$, $x_{B}$, $\tilde{x}_{A}$, and $\tilde{x}_{B}$.
The first and second items in equation~\ref{eq:loss_quad} are aim to minimize intra-class variation and maximizing means of inter-classes.
The final term acts differently, it aims to auxiliary regulate the inter-class means.

\subsubsection{Overall training objective}
The training objective of a classifier $C$ in GTCN is:
\small
\begin{equation}\label{eq:loss_gen_all}
L_{C} = L_{cls} + L_{quad}~.
\end{equation}
\normalsize
The training objective of translator $G$ in GTCN is:
\small
\begin{equation}\label{eq:loss_gen_all}
L_{G} =  L_{C} + L_{adv}^{A} + L_{adv}^{B} + \lambda \cdot (L_{cyc}^{A} + L_{cyc}^{B})~,
\end{equation}
\normalsize
where $\lambda$ is a weight parameter to adjust the relative importance of the cyclic consistency losses.
Finally, we aim to optimize:
\small
\begin{align}\label{eq:final_optimization}
G_{AB}^{*}, G_{BA}^{*} & = \arg \underset{\{G_{AB},G_{BA}\}} \min \underset{\{D_{A},D_{B}\}} \max L_{G}~,
\end{align}
\begin{align}\label{eq:final_optimization}
C^{*} &= \arg \underset{\{C\}} \min~L_{C}~.
\end{align}
\normalsize
Since the translators $G_{AB}$, $G_{BA}$ and the discriminators $D_{A}$, $D_{B}$ are jointly optimized with $C$,
$G_{AB}$ and $G_{BA}$ generate effective translated samples that assist improvement of classification accuracy for $C$.

\subsection{Networks Model}
As the baseline translator network, we adapt the implementation of CycleGAN. \footnote{\url{https://junyanz.github.io/CycleGAN/}}.
All of the classifiers in the experiments use the simple architecture consisting of six convolution layers, so as to run on performance-critical systems such as smartphones.
The parameters are denoted:
\small
\begin{align}\label{eq:set_classifier}
\theta_{C} = \{w^{1}, w^{2}, w^{3}, w^{4}, w^{5}, w^{6}, w^{s}\}.
\end{align}
\normalsize
The specific parameters for the network are given in Table ~\ref{tab:classifier_define}.
As $\theta_{C}$ only consists of 73,904~($k$=2)~/~75,952~($k$=4) parameters,
the trained networks $C$s have a small memory footprint and can be run on smart devices in real-time without GPU acceleration.
\begin{table}[!t] 
  \caption{Design of a light-weight classifier $C$. Resolution of input image $x$ is $128\times128$.
  $Conv$ stands for a convolutional layer, $BN$ for batch normalization, $ReLU$ for rectified linear units, and $Pool$ corresponds to a pooling layer.
  $FC$ denotes a fully connected layer.
  $k$ is the number of classes for $y\in Y$.}
  \label{tab:classifier_define}
  \centering
  \small
  \tabcolsep=0.16cm
  \scalebox{0.9}{
  \begin{tabular}{ccc}
  \toprule
  Layers & Parameters & Outputs\\
  \midrule
  $Conv1$-$BN$-$ReLU$       & $\it{w^{1}}$: $3\times3\times3$, kernels=16  & $128\times128\times16$ \\
  $Pool1$                   & max: $2\times2$, stride=2                    & $64\times64\times16$ \\
  \midrule
  $Conv2$-$BN$-$ReLU$        & $\it{w^{2}}$: $3\times3\times16$, kernels=16 & $64\times64\times16$ \\
  $Pool2$                   & max: $2\times2$, stride=2                    & $32\times32\times16$ \\
  \midrule
  $Conv3$-$BN$-$ReLU$       & $\it{w^{3}}$: $3\times3\times16$, kernels=32 & $32\times32\times32$ \\
  $Pool3$                   & max: $2\times2$, stride=2                    & $16\times16\times32$ \\
  \midrule
  $Conv4$-$BN$-$ReLU$       & $\it{w^{4}}$: $3\times3\times32$, kernels=32 & $16\times16\times32$ \\
  $Pool4$                   & max: $2\times2$, stride=2                    & $8\times8\times32$ \\
  \midrule
  $Conv5$-$BN$-$ReLU$       & $\it{w^{5}}$: $3\times3\times32$, kernels=64 & $8\times8\times64$ \\
  $Conv6$-$BN$-$ReLU$       & $\it{w^{6}}$: $3\times3\times64$, kernels=64 & $8\times8\times64$ \\
  \midrule
  $Pool6$                   & average: $2\times2$, stride=2                & $4\times4\times64$ \\
  $Dropout$                 &  -                                           & $1024$\\
  \midrule
  $FC$                      & $\it{w^{s}}$:$1024\times k$                  & $k$\\
  $Softmax$                 & -                                            & $P(y|x)$\\
  \bottomrule
  \end{tabular}
  }
  \normalsize
\end{table}

\section{Experiments}\label{sec:experiments}
\subsection{Experimental Setup and Configuration}
\begin{figure}[!t] 
  \centering
  \centerline{\includegraphics[width=6cm]{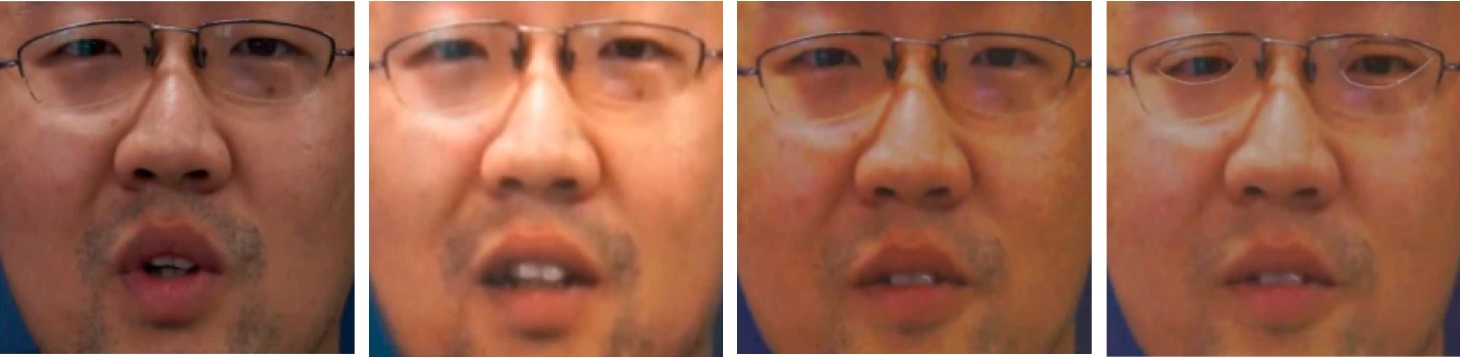}}
  \centerline{\includegraphics[width=6cm]{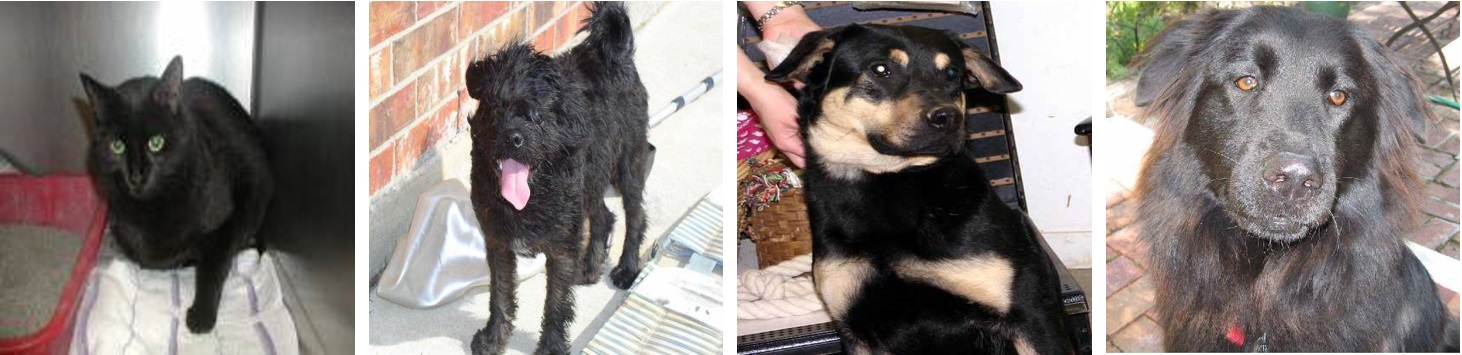}}
  \caption{Example from face liveness and dogs vs. cats.
           (First row) The first column shows real face images. The second, third, and fourth column show fake face images those are corresponding to video display, printed photos, and face masks with real eyes.
           (Second row) The first column shows cat images, others show dog images.}
  \label{fig:DATASET_EXAMPLE}
\end{figure}

The proposed methods are evaluated with datasets,
those are (1) \textit{Face liveness} and (2) \textit{Dogs vs. Cats} for visually similar binary classes.
The datasets represent human faces and animals, and as a result have very different characteristics in terms of underlying structure and styles.
The face liveness detection dataset\footnote{\url{http://www.cbsr.ia.ac.cn/english/FASDB_Agreement/Agreement.pdf}}\cite{zhang2012face} contains very similar fake and real face images that have strong underlying structure and different style of texture.
Note that no subjects are present in both training and test sets.
Additionally, we evaluate on the dogs vs. cats classification dataset\footnote{\url{https://www.kaggle.com/c/dogs-vs-cats}}, because they share part of similar style although they are belong to the different species.
We explore varying the amount of training data available by training on different subsets containing 100\%, 80\%, 60\%, and 40\% of the examples of the original training set.
The specific configuration is described in Table ~\ref{tab:db_define_binary} and examples images of datasets are shown in Figure ~\ref{fig:DATASET_EXAMPLE}.

We trained all the models in the experiments from scratch without pre-training and extra datasets.
The mini-batches used to train the CNNs contained eight real samples, while four real samples and four translated samples were present in the mini-batchs used for GTCNs and other compared models.
All of the networks' parameters were optimized using the Adam optimizer.
We do not vary the learning rate for the first half of epochs and linearly decay the rate to zero over the next half of epochs.
The base learning rate is 0.0002 and the number of training epochs was set to 100.
To prevent over-fitting, data augmentation transforms were applied such as rotation, intensity, color adjustment, and scaling variation.
Regarding the hyper parameters of the translator network, $\lambda$ was set to 10 in all experiments.
The hyperparameters for the quadruplet loss were set to: \{$\eta_{a}/\eta_{b}$=2, $\eta_{c}$=6\} for face liveness, \{$\eta_{a}/\eta_{b}$=0.5, $\eta_{c}$=8\} for dogs vs. cats.

Since binary classification is the problem of deciding to which class $y \in Y=\{A,B\}$ a given test image $x$ belongs,
we utilize logit values obtained before calculating the Softmax instead of using the Softmax output that was used for training.
Outputs of the $FC$ layer in the classifier $C$ are utilized to calculate a score for class $A$ as:
$SC(y=A|x;\theta_{C})=\frac{FC_{A}-FC_{B}}{2}~$,
where $FC_{A}$ is a logit value for class $A$, $FC_{B}$ is a logit value for class $B$, $SC(y=A)$ is the score for class $A$.
The calculated $SC(y=A)$ is employed to decide on the class as follows:
\small
\begin{align}\label{eq:binary_classification}
C(x;\theta_{C}) = \begin{cases}
y=A &\text{if $SC(y=A)\geq th$}\\
y=B &\text{otherwise}
\end{cases}~,
\end{align}
\normalsize
where $th$ is an acceptance threshold to decide if $x$ is a class $A$.
For example, if $th$ is set to be high, then we can calculate TAR for overall test samples at low FAR.

\begin{table}[!t] 
  \caption{Configuration of experimental datasets and subsampled variants.
           LV{*} and DC{*} are reconfigured training datasets.}
  \label{tab:db_define_binary}
  \centering
  \small
  \tabcolsep=0.18cm
  \scalebox{0.9}{
  \begin{tabular}{lccccc}
    \toprule
    Face liveness       & LV100    & LV80     & LV60     & LV40    & Test    \\
    \midrule
    Number of subjects  & 20       &  16      & 12       & 8       & 30      \\
    Live face           & 10,891   &  8,333   & 6,443    & 4,493   & 15,904  \\
    Fake face           & 34,165   &  26,557  & 20,188   & 13,511  & 49,862  \\
    \toprule
    Dogs vs. cats       & DC100    & DC80     & DC60    & DC40    & Test   \\
    \midrule
    Dogs                & 12,500   &  10,000  & 7,500    & 5,000   & 6,253  \\
    Cats                & 12,500   &  10,000  & 7,500    & 5,000   & 6,235  \\
    \bottomrule

  \end{tabular}
  }
  \normalsize
\end{table}

\subsection{Training with small data volumes} \label{sec:small_dataset}
First, we evaluate performance of baseline CNNs on different subsets of the training set to study the  effect of data scarcity.
In addition to this, other compared methods and the proposed GTCN are trained with only 40\% of training data.
We compared our method to between-class learning (BC/BC$^{+}$)\cite{tokozume2017between} and semi-supervised learning with the classical $N+1$ class setting~\cite{salimans2016improved}.
In terms of generative models, we consider a least-squares GAN (LSGAN)\cite{mao2017least} and a variational autoencoder (VAE)\cite{kingma2013auto} as alternative methods to be compared.
Table \ref{tab:CASIA_db_eval_cnn} and Table \ref{tab:CATDOG_db_eval_cnn} show evaluation results.
As the two binary classes are visually similar and there is therefore a lack of diverse data, baseline CNNs have low accuracy despite using deep networks and 100\% of training data in both datasets.
The overall sparsity of training data causes a poor true acceptance rate. 
Interestingly, BC/BC$^{+}$ did not fare better than CNNs in the range of low false acceptance rates, although mean accuracy for those methods was higher than for CNNs.
We hypothesize this is due to the fact that BC/BC$^{+}$ may be hard to produce good mixing data in cases where classes are very similar, caused by the proximity of the manifolds corresponding to those classes.
All deep generative models outperformed the CNN baseline for LV40 dataset.
In particular, VAE-based methods achieve a good accuracy, most likely because such methods generate diverse samples.
For all evaluation datasets, GTCNs that were trained with 40\% of the dataset clearly outperformed all of other compared methods including CNNs trained on 100\% of training data.
With GTCNs that were trained with 20\% of the dataset, this results in lower accuracy compared to CNNs trained on 100\% of training data.
As a result, we consider that a dataset comprising 40\% of the examples is an empirical lower-bound on “acceptable” performance in this context.
\begin{table}[!t] 
  \caption{Evaluation results of training with small volume of the face liveness dataset.
   ACC is mean accuracy. Cells in columns of FAR show percentage value of TAR.}
  \label{tab:CASIA_db_eval_cnn}
  \centering
  \small
  \tabcolsep=0.07cm
  \scalebox{0.9}{
  \begin{tabular}{llccccc}
    \toprule
    Model               & Dataset   & ACC    & FAR=$\frac{1}{100}$ & FAR=$\frac{1}{1k}$ & FAR=$\frac{1}{5k}$ & FAR=$\frac{1}{50k}$ \\
    \midrule
    CNN                 & LV40      & 74.71  & 43.42  & 22.52 & 16.90 &  9.75 \\
    CNN                 & LV60      & 81.16  & 50.26  & 27.85 & 20.82 & 13.20 \\
    CNN                 & LV80      & 86.79  & 64.13  & 41.38 & 27.77 & 22.29 \\
    CNN                 & LV100     & 84.87  & 69.33  & 45.38 & 32.77 & 21.72 \\
    \midrule
    BC                  & LV40      & 88.62  & 43.13  & 17.78 & 12.61 & 5.28  \\
    BC$^{+}$            & LV40      & 86.09  & 28.10  & 11.78 &  7.65 & 5.48  \\
    LSGAN       & LV40      & 79.62  & 60.88  & 37.47 & 30.25 & 23.58 \\
    VAE         & LV40      & 90.57  & 67.93  & 42.71 & 32.94 & 17.54 \\
    Semi-sup.    & LV40      & 89.89  & 58.49  & 41.95 & 33.63 & 21.64 \\
    \midrule
    GTCN                & LV40      & \bfseries92.26  & \bfseries74.81  & \bfseries62.36 & \bfseries54.04 & \bfseries39.01 \\
    \bottomrule
  \end{tabular}
  }
  \normalsize
\end{table}

\begin{table}[!t] 
  \caption{Evaluation results of training with small volume of the dogs vs. cats dataset.
           EER is the equal error rate.}
  \label{tab:CATDOG_db_eval_cnn}
  \centering
  \small
  \tabcolsep=0.12cm
  \scalebox{0.9}{
  \begin{tabular}{llccccc} 
    \toprule
     Model          &Dataset   & ACC     & FAR=$\frac{1}{100}$ & FAR=$\frac{1}{1k}$ & FAR=$\frac{1}{5k}$ & EER \\
    \midrule
    CNN             &DC40      & 92.74   & 79.49  & 54.82 &37.26  & 7.50 \\
    CNN             &DC60      & 93.38   & 81.92  & 65.71 &44.03  & 6.60 \\
    CNN             &DC80      & 93.63   & 82.90  & 66.43 &42.85  & 6.25 \\
    CNN             &DC100     & 93.96   & 82.98  & 63.96 &58.83  & 6.06 \\
    \midrule
    BC              &DC40      & 92.82   & 77.13  & 28.77 & 5.16  & 7.23 \\
    BC$^{+}$        &DC40      & 92.33   & 69.94  & 23.08 & 4.78  & 8.14 \\
    LSGAN           &DC40      & 93.13   & 79.44  & 54.88 & 25.07 & 6.84 \\
    VAE             &DC40      & 92.89   & 79.10  & 56.97 & 50.97 & 7.12 \\
    Semi-sup.       &DC40      & 92.67   & 76.26  & 57.64 & 37.21 & 7.18 \\
    \midrule
    GTCN            &DC40      & \bfseries94.28   & \bfseries84.15  & \bfseries67.41 & \bfseries54.18& \bfseries5.66 \\
    \bottomrule
  \end{tabular}
  }
  \normalsize
\end{table}

\begin{table}[!t] 
  \caption{Evaluation results of training with full volume of the face liveness dataset(LV100).
          All of models use images of 128$\times$128 pixels, except the CNN-256, which uses 256$\times$256. $^{*}$ are results of score fusion based model.}
  \label{tab:CASIA_GACN_results_full}
  \centering
  \small
  \tabcolsep=0.15cm
  \scalebox{0.9}{
  \begin{tabular}{lccccc}
    \toprule
    Model             & ACC & FAR=$\frac{1}{100}$ & FAR=$\frac{1}{1k}$ & FAR=$\frac{1}{5k}$ & FAR=$\frac{1}{50k}$ \\
    \midrule
    CNN               & 84.87    & 69.33  & 45.38 & 32.77 & 21.72  \\
    CNN-256           & 91.09    & 81.68  & 64.51 & 58.87 & 48.16  \\
    \midrule
    BC                & 90.17    & 56.93  & 23.84 & 16.53 & 10.92  \\
    BC$^{+}$          & 91.19    & 52.97  & 29.28 & 21.20 & 8.51  \\
    LSGAN             & 93.88    & 81.21  & 50.97 & 37.19 & 26.24 \\
    VAE               & 95.50    & 86.71  & 77.78 & 70.09 & 61.63 \\
    Semi-sup.         & 94.65    & 76.75  & 43.26 & 31.18 & 18.04 \\
    \midrule
    GTCN              & \bfseries97.65    & \bfseries93.62  & \bfseries86.74 & \bfseries81.81 & \bfseries76.73 \\
    \bottomrule
    \toprule
    CNN$^{*}$         & 95.76    & 93.15  & 82.12 & 79.41 & 61.94  \\
    CNN-256$^{*}$     & 97.69    & 96.34  & 91.57 & 83.62 & 74.16  \\
    \midrule
    GTCN$^{*}$        & \bfseries99.03    & \bfseries98.87  & \bfseries95.99 & \bfseries90.73 & \bfseries85.76  \\
    \bottomrule
  \end{tabular}
  }
  \normalsize
\end{table}

\begin{table}[!t] 
  \caption{Evaluation results of training with full volume of the dogs vs. cats dataset(DC100). CNN-256 uses 256$\times$256.} %
  \label{tab:CATDOG_GACN_results_full}
  \centering
  \small
  \tabcolsep=0.21cm
  \scalebox{0.9}{
  \begin{tabular}{lccccc}
    \toprule
    Model          & ACC       & FAR=$\frac{1}{100}$ & FAR=$\frac{1}{1k}$ & FAR=$\frac{1}{5k}$ & EER \\
    \midrule
    CNN            & 93.96     & 82.98   & 63.96  & 58.83 & 6.06 \\
    CNN-256        & 95.52     & 89.83   & 78.85  & 69.03 & 4.51 \\
    \midrule
    BC             & 93.79     & 78.46   & 32.98  & 17.93 & 6.49 \\
    BC$^{+}$       & 93.68     & 83.80   & 38.83  & 19.44 & 7.00 \\
    LSGAN          & 95.21     & 86.09   & 72.46  & 58.41 & 4.81 \\
    VAE            & 94.58     & 84.88   & 68.24  & 60.14 & 5.49 \\
    Semi-sup.      & 94.86     & 85.32   & 67.30  & 59.50 & 5.18 \\
    \midrule
    GTCN           & \bfseries95.61	   & \bfseries88.44   & \bfseries75.22  & \bfseries66.72 & \bfseries4.37 \\
    \bottomrule
  \end{tabular}
  }
  \normalsize
\end{table}

\begin{table}[!t] 
  \caption{Recall results of training with full volume of the face liveness dataset (LV100) for sub-categories}
  \label{tab:CASIA_GACN_specific_results}
  \centering
  \small
  \tabcolsep=0.23cm
  \scalebox{0.9}{
  \begin{tabular}{lcccc}
    \toprule
    Model        & Real Face & Video Face & Photo Face & Mask Face\\
    \midrule
    CNN          & 96.28    & 83.05  & 73.51  & 89.35  \\
    GTCN         & 93.68    & 97.29  & 98.04  & 99.98  \\
    \bottomrule
  \end{tabular}
  }
  \normalsize
\end{table}

\begin{table}[!ht] 
  \caption{Performance comparisons with the deep learning based face liveness detection methods}
  \label{tab:CASIA_SOTA_COMPARISON}
  \centering
  \small
  \tabcolsep=0.15cm
  \scalebox{0.9}{
  \begin{tabular}{lcccl}
    \toprule
    Method                            & EER          & Image Resolution  & Methods                 \\
    \toprule
    Yang\cite{yang2014learn}          & 4.95         & 128$\times$128    & AlexNet + SVM           \\
    DPCNN\cite{li2016original}        & 4.50         & 224$\times$224    & VGG19 + SVM             \\
    Atoum\cite{atoum2017face}         & 2.67         & 128$\times$128    & CNN (10 patch + depth)  \\
    Nguyen\cite{nguyen2018combining}  & 1.70         & 224$\times$224    & VGG19 + MLBP +SVM       \\
    \midrule
    CNN$^{*}$                         & 3.78         & 128$\times$128    & CNN (Face + Context)    \\
    CNN-256$^{*}$                     & 2.53         & 256$\times$256    & CNN (Face + Context)    \\
    \midrule
    GTCN                              & 3.49         & 128$\times$128    & CNN (Face)              \\
    GTCN                              & 2.09         & 128$\times$128    & CNN (Context)           \\
    GTCN$^{*}$                        &\bfseries1.02 & 128$\times$128    & CNN (Face + Context)    \\
    \bottomrule
  \end{tabular}
  }
  \normalsize
\end{table}

\begin{table}[!ht] 
  \caption{Comparison of architectures in term of number of parameters, number of layers, MACs (Multiply-Accumulate Operation), and latency on Mobile Application Processor}
  \label{tab:MODEL_COMPLEX_COMPARISON}
  \centering
  \small
  \tabcolsep=0.07cm
  \scalebox{0.9}{
  \begin{tabular}{lcccr}
    \toprule
    Base Model                                       & Parameters & Layers & MACs    & Speed @Mobile AP   \\
    \toprule
    AlexNet\cite{yang2014learn}                      & 61M        & 8      & 181M    & 100ms @CPU           \\
    VGG19\cite{nguyen2018combining, li2016original}  & 143M       & 19     & 19.7G   & 170ms @GPU          \\
    Light-weight $C$                                 & 74K        & 6      & 27M     & 15ms @CPU          \\
    \bottomrule
  \end{tabular}
  }
  \normalsize
\end{table}

\subsection{Training with the full dataset}
\begin{figure}[!t]
\centering
\includegraphics[height=1.8cm]{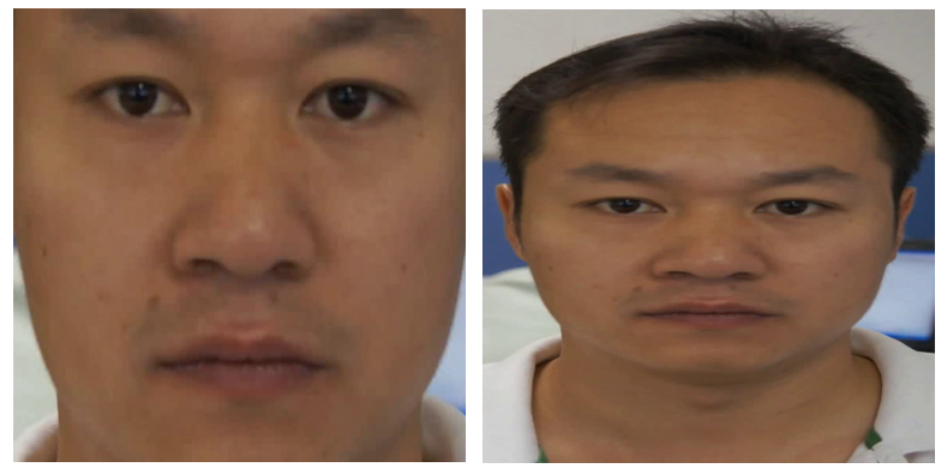}
\caption{Examples of image patches: resized detected face $x_{rs}$, and resized contextual $x_{ct}$ from input image $I$. 
Note that $x_{rs}$ and $x_{ct}$ are images of $128\times128$ pixels.} 
\label{fig:patches}
\end{figure}

In this experiment, we used 100\% of the training dataset to verify the scalability of the proposed methods.
We added CNN-256 models that are trained and tested on images of size 256$\times$256 pixels, while other models use images of size 128$\times$128 pixels.
Table \ref{tab:CASIA_GACN_results_full} and Table \ref{tab:CATDOG_GACN_results_full} show evaluation results.
In a similar fashion to the experiments using reduced versions of the datasets, the performance of data augmentation methods BC/BC$^{+}$ and the compared deep generative models that mainly employ intra-class data augmentation show limited performance.
VAEs and BC methods show relatively good performance in the face liveness dataset, since faces are highly structured images.
However, in both evaluation datasets, GTCNs outperform all of other compared methods including CNNs trained with a larger input, 256$\times$256 pixels.
As a matter of course, a small size of input can lead to two advantages, 1) light-weight inference and 2) recognizing far-distance images.
Effectiveness of GTCNs for each sub-category in the face liveness dataset such as real face, video face, photo face, and mask face is shown in Table \ref{tab:CASIA_GACN_specific_results}.
Recall=$\frac{\text{True Positives}}{\text{True Positives} + \text{False Negatives}}$ is chosen as an evaluation metric.
In sub-category of fake faces, the recall is clearly improved, while the recall of real faces is negligibly degraded.
Lastly, we use score fusion models to achieve the best accuracy in the face liveness detection.
Formally, we consider two classifiers, $C_{rs}$ and $C_{ct}$.
The classifier $C_{rs}$ that discern local texture, shape and reflection pattern is given by
\begin{equation}\label{eq:eq1}
SC_{rs}=C_{rs}(x_{rs};\theta_{rs}), \forall x_{rs} \in Resize(FD(I)),
\end{equation}
where $SC_{rs}$ is a liveness score of a patch $x_{rs}$ that is a resized image of a detected face region,
$FD$ is a face detector \cite{viola2001rapid}, and $\theta_{rs}$ are learned parameters.
Similarly, the classifier $C_{ct}$ that discriminates global shapes and contextual pattern is given by
\begin{equation}\label{eq:eq3}
SC_{ct}=C_{ct}(x_{ct};\theta_{ct}), \forall x_{ct} \in Resize(I),
\end{equation}
where $SC_{ct}$ is a liveness score of a patch $x_{ct}$ that is a resized contextual image from an input image $I$, and $\theta_{ct}$ are learned parameters.
To decide whether the given image $I$ is live or not, the final liveness score is calculated by
\begin{equation}\label{eq:eq4}
SC = \alpha_{rs} \times SC_{rs} + \alpha_{ct} \times SC_{ct},
\end{equation}
where $\it{\alpha_{rs}}$ and $\it{\alpha_{ct}}$ are coefficient values to combine scores from two models to perform liveness prediction with a late fusion method.
We set $\it{\alpha_{rs}}$=1 and $\it{\alpha_{ct}}$=0.6, respectively.
Example of patches are shown in Figure \ref{fig:patches}.

\begin{figure}[!t] 
  \centering
  \centerline{
    {\includegraphics[width=4.2cm]{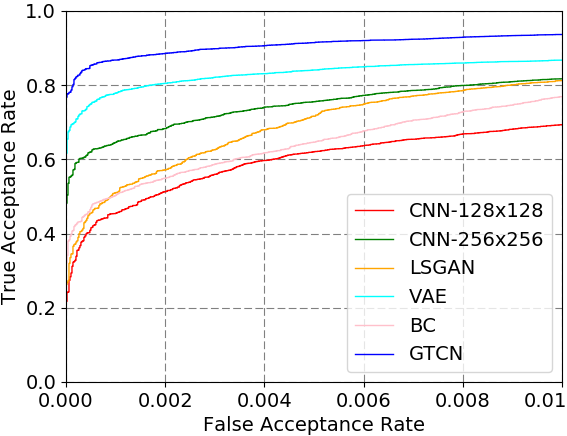}} %
    \hspace*{-0.18cm}
    {\includegraphics[width=4.2cm]{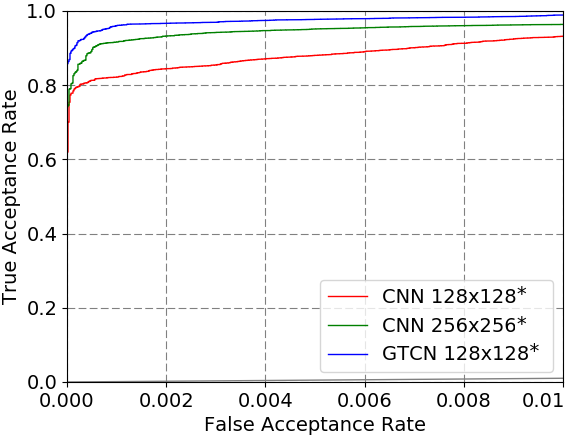}} %
  }
  \caption{Receiver Operating Characteristic (ROC) comparison of models trained with full volume of the face liveness dataset.
  (Left) ROC curves of single models
  (Right) ROC curves of two patches based models}
  \label{fig:experiment_full}
\end{figure}

The proposed methods outperform previous state-of-the-art models and score fusion based models of CNN and CNN-256 as shown in table \ref{tab:CASIA_SOTA_COMPARISON} without pretraining with extra datasets, large networks, and combining SVM with traditional hand craft features.
Figure \ref{fig:experiment_full} shows the ROC curves for single models and two patches based models.
GTCNs achieve superior true acceptance rates in the range of low false acceptance rates.
Note that our two patches based model is lighter and faster than others, since we used a light-weight CNN that has just 73,904 parameters.
In Table \ref{tab:MODEL_COMPLEX_COMPARISON}, there are specific comparison with the compared methods that employ other CNN architecture such as AlexNet and VGG19, while our proposed architecture consists of just 6 convolution layers.
In our implementation, the ensemble of the two light-weight networks for two facial patches is running in 30 msec by using single mobile CPU on Samsung Galaxy S10.

\begin{table}[!t] 
  \caption{Ablation experimental results of the proposed methods on face liveness and dogs vs. cats datasets}
  \label{tab:ablation_study}
  \centering
  \small
  \tabcolsep=0.1cm
  \scalebox{0.9}{
  \begin{tabular}{lcccccc} 
      \toprule
      Face liveness          & CNN   & Separate & +Joint & +AF       & +AF/QL   & +AF/QL/ST \\
      \midrule
      Accuracy               & 84.87 & 94.93 & 96.43   & 95.35    & 97.04 & \bfseries97.65  \\
      FAR=$\frac{1}{50k}$    & 21.72 & 27.15 & 67.10   & 55.71    & 74.42 & \bfseries76.73  \\
      \bottomrule
      \toprule
      Dogs vs. cats        & CNN & Separate & +Joint    & +AF     & +AF/QL   & +AF/QL/ST \\
      \midrule
      Accuracy            & 93.96 & 95.05 &95.44 & 95.72  & 94.93 & \bfseries95.61  \\
      FAR=$\frac{1}{5k}$  & 58.83 & 42.25 &64.38 &57.37  & 60.29 & \bfseries66.72  \\
      \bottomrule
  \end{tabular}
  }
 \normalsize
\end{table}

\begin{figure}[!t] 
  \centering
  \centerline{\includegraphics[width=7.0cm]{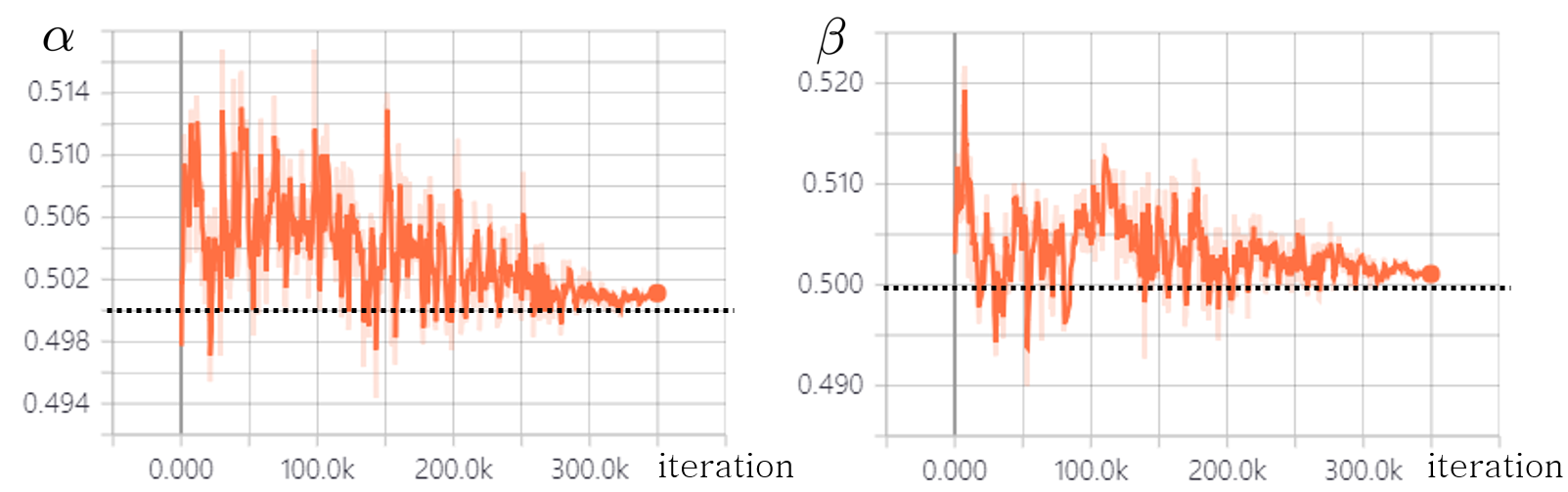}}
  \caption{Example variation of $\alpha$ and $\beta$ for adaptive fade-in learning}
  \label{fig:experiment_AF}
\end{figure}

\begin{figure}[!t]
\centering
\centerline{
\includegraphics[width=4.0cm]{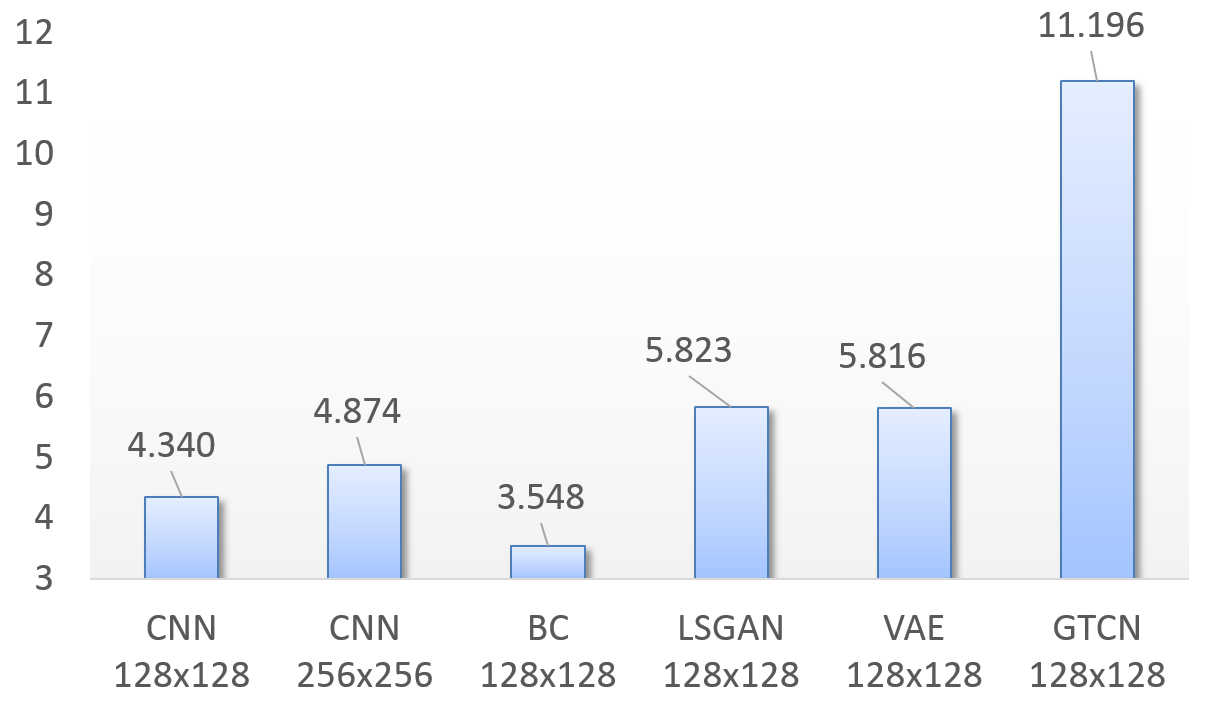}
\includegraphics[width=4.0cm]{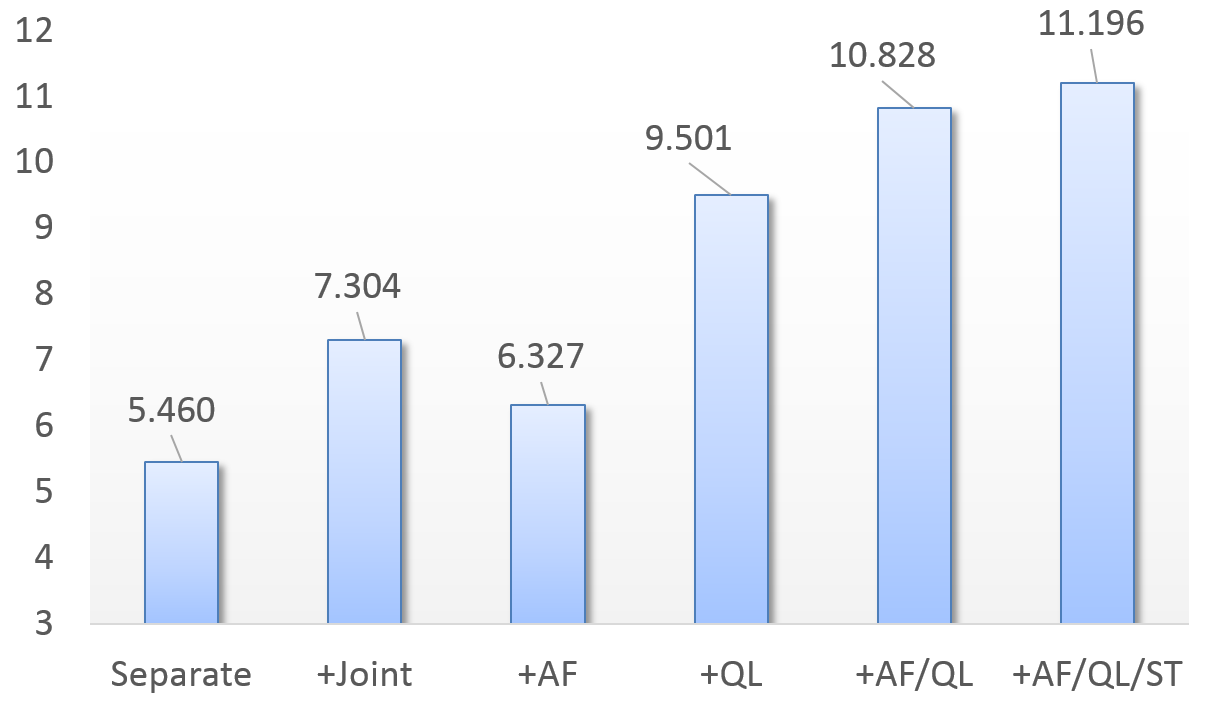}
}
\caption{Comparison of Fisher's criterion scores. Vertical axis on the charts represents score $J$. (Left) Compared methods (Right) Ablation study results of the proposed GTCN}
\label{fig:FISHER_SCORE_LV}
\end{figure}

\begin{figure}[!t]
\centering
\centerline{
    \hspace{1mm}
    \includegraphics[width=3.2cm]{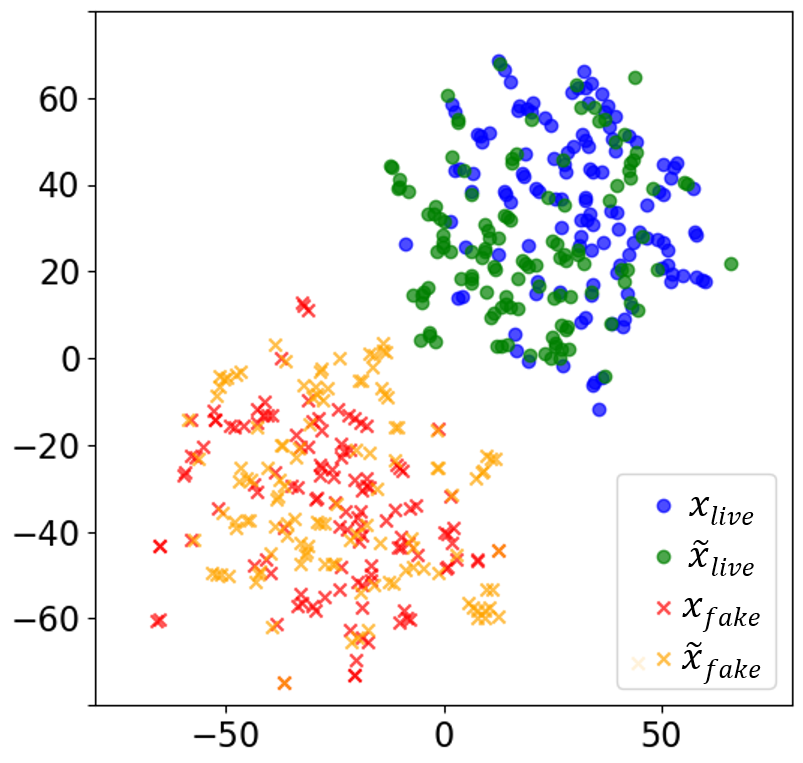}
    \hspace{0.5mm}
    \includegraphics[width=3.2cm]{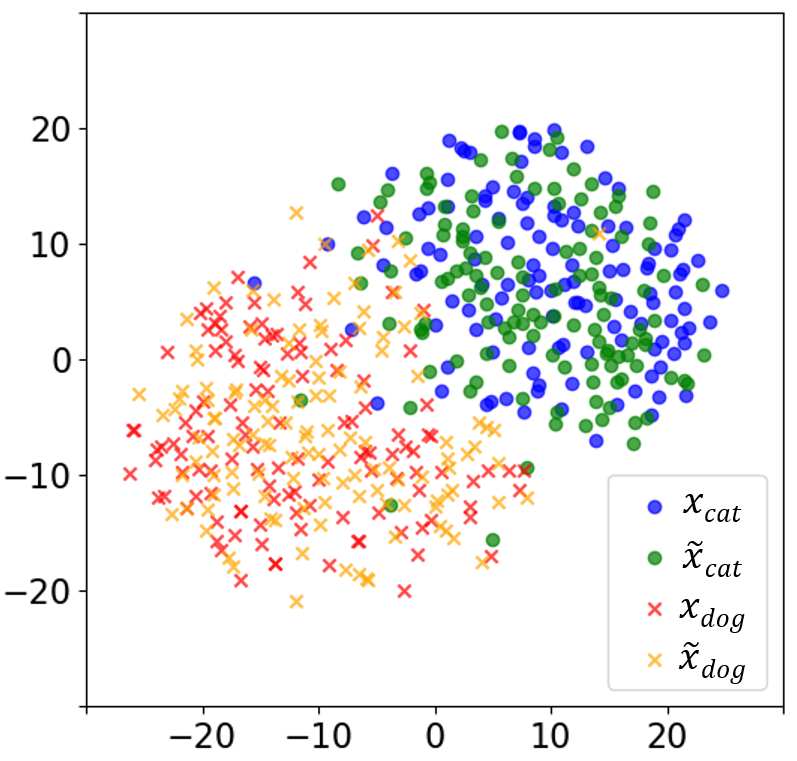}
    }
\centerline{
    \includegraphics[width=3.2cm]{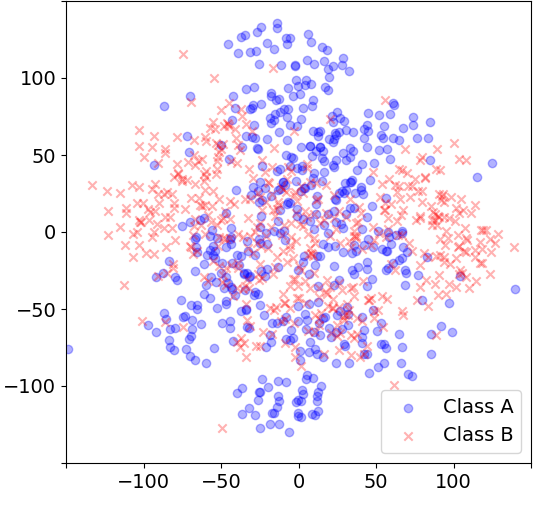}%
    \hspace{+0.7mm}
    \includegraphics[width=3.2cm]{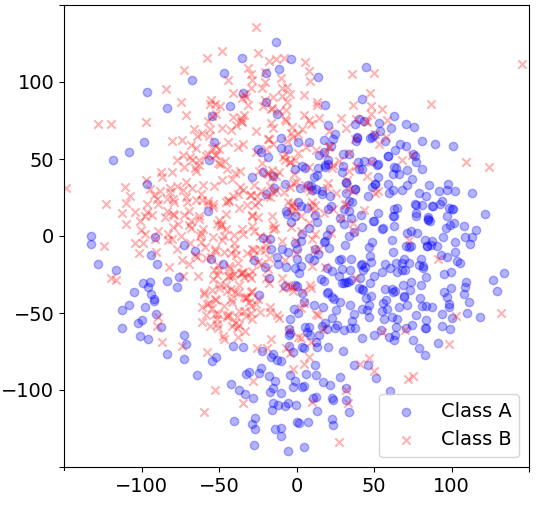}%
  }
\caption{
First two plots show examples of t-SNE analysis for training data on GTCN:
(1) Face liveness and (2) Dogs vs. cats.
The augmented data points $\tilde{x}$ achieve a good coverage of all the area of the embedding space corresponding to their true class in binary classification.
We can consider that they are good at augmenting training data in that regard.
Second two plots show comparison of methods by t-SNE analysis for LV40 test dataset:
(3) CNN and (4) the proposed GTCN
}
\label{fig:TSNE_TRAINING}
\end{figure}

\section{Discussion and Analysis}
\subsection{Ablation Study}
We perform an ablation study, the results of which are shown in Table \ref{tab:ablation_study}.
As first, we trained $G_{AB}$ and $G_{BA}$ separately and used translators with fixed parameters to augment data for training $C$.
$Seperate$ learning shows better accuracy then baseline CNN, because translators generate diverse data.
However, proposed $Joint$ learning is superior to $Separate$ learning.
One of the disadvantages of $Separate$ learning is that it takes longer to train, since two models are trained sequentially.
To analyze the effect of each component of the GTCN, learning options were adapted sequentially.
AF stands for adaptive fade-in learning, QL for quadruplet loss, and ST corresponds to stochastic translation.
Because translated images could be sometimes suppressed by applying AF during joint training, the performance are not much improved on binary classification.
Figure ~\ref{fig:experiment_AF} shows the example values taken by $\alpha$ and $\beta$ for the AF method during training of the face liveness dataset.
At the start of training, $\alpha$ and $\beta$ were higher than 0.5, because real images are more confidently classified than generated images, while the values have almost converged to 0.5 at the end of training in the example.
However, when AF, QL and ST are applied simultaneously, the performance of GTCNs is fairly improved,
since ST contributes data diversity, QL increases fisher's criterion, and AF controls noisy images.
\begin{figure*}[!t]
\centering
\centerline{
    \includegraphics[width=6cm]{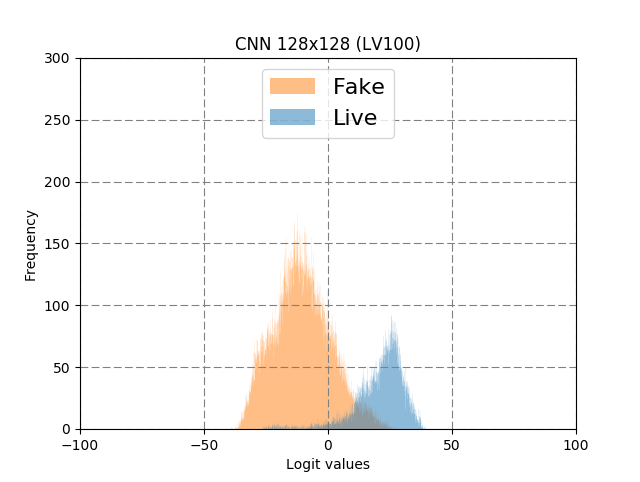}
    \hspace{-2mm}
    \includegraphics[width=6cm]{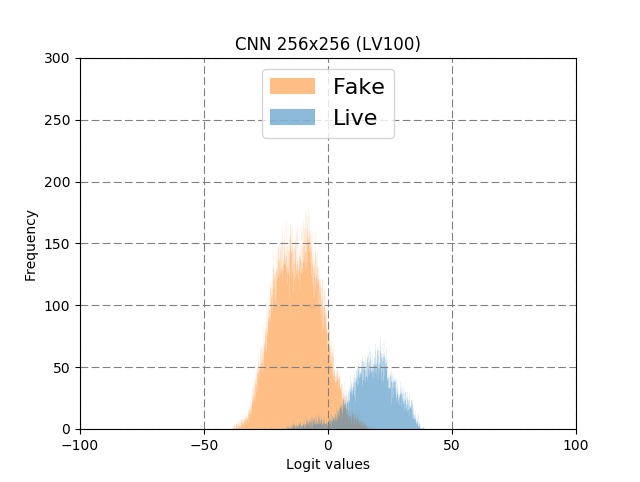}
    \hspace{-2mm}
    \includegraphics[width=6cm]{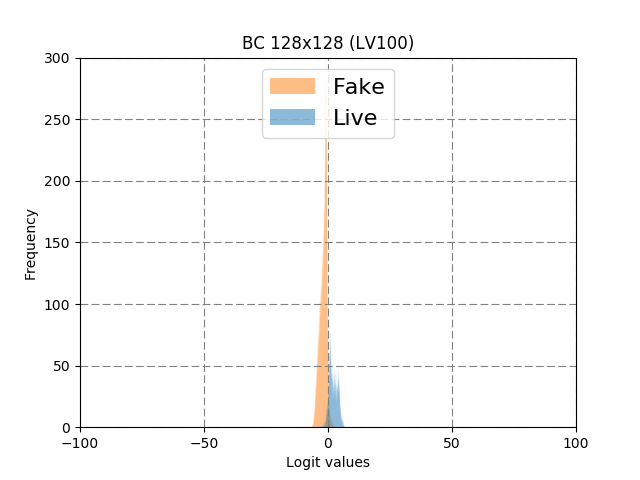}
    \hspace{-2mm}
}
\centerline{
    \includegraphics[width=6cm]{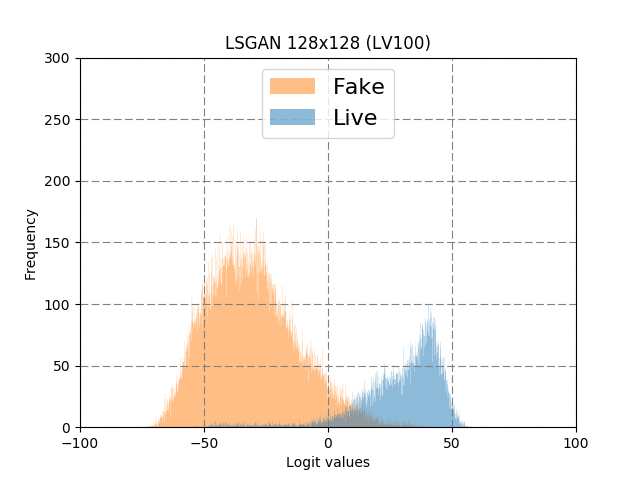}
    \hspace{-2mm}
    \includegraphics[width=6cm]{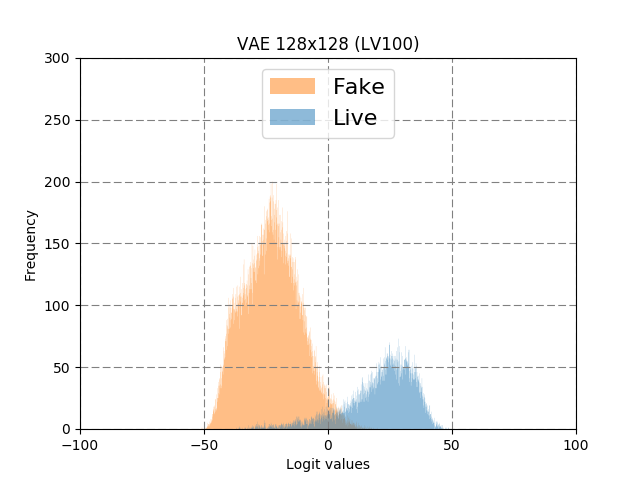}
    \hspace{-2mm}
    \includegraphics[width=6cm]{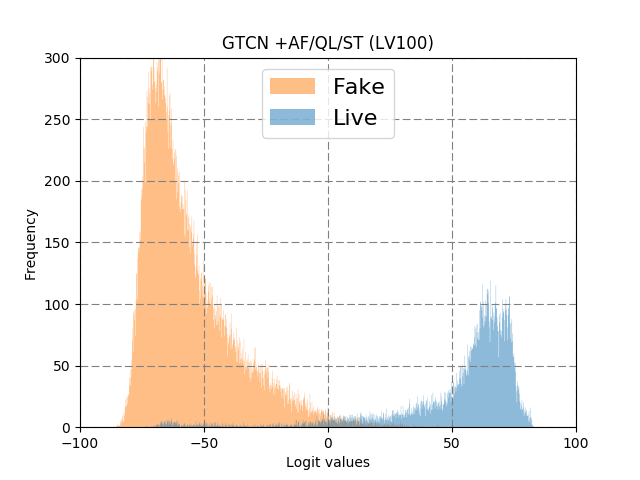}
    \hspace{-2mm}
}
\caption{Histograms of logit values for liveness test dataset. The compared models are trained on LV100 dataset. (1) CNN 128$\times$128 (2) CNN 256$\times$256 (3) BC 128$\times$128 (4) LSGAN 128$\times$128 (5) VAE 128$\times$128 (6) GTCN 128$\times$128}
\label{fig:histogram_methods_liveness1}
\centerline{
    \includegraphics[width=6cm]{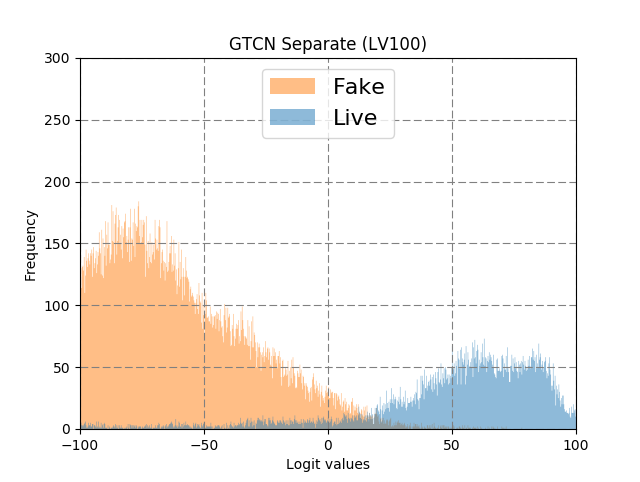}
    \hspace{-2mm}
    \includegraphics[width=6cm]{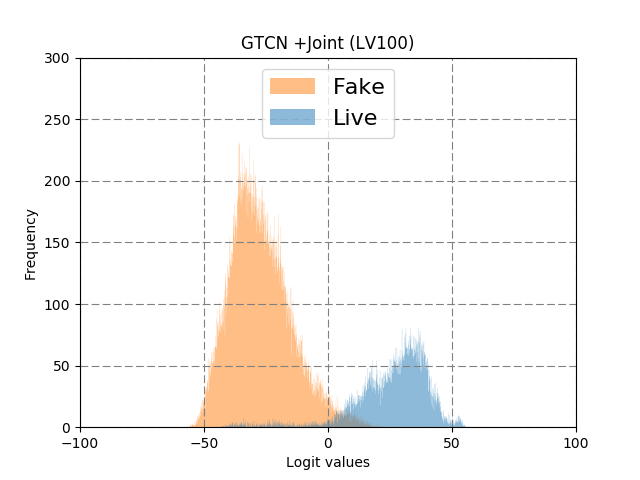}
    \hspace{-2mm}
    \includegraphics[width=6cm]{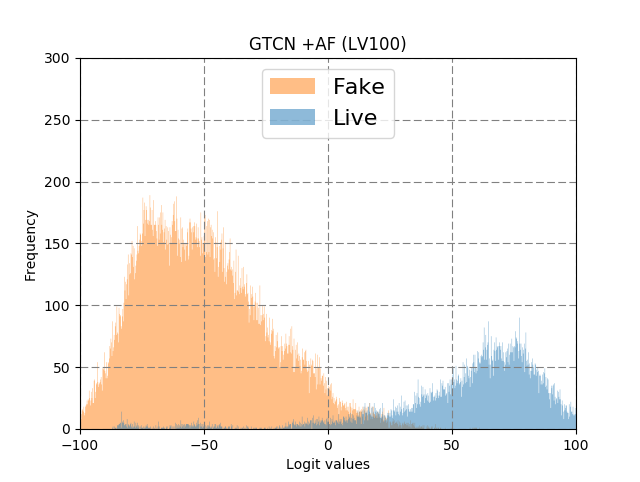}
    \hspace{-2mm}
}
\centerline{
    \includegraphics[width=6cm]{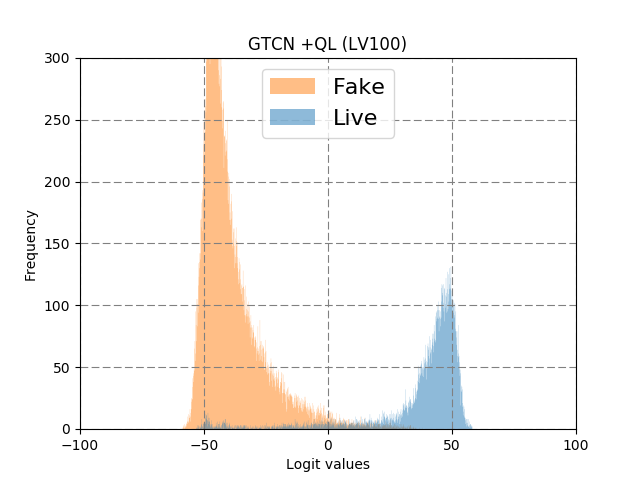}
    \hspace{-2mm}
    \includegraphics[width=6cm]{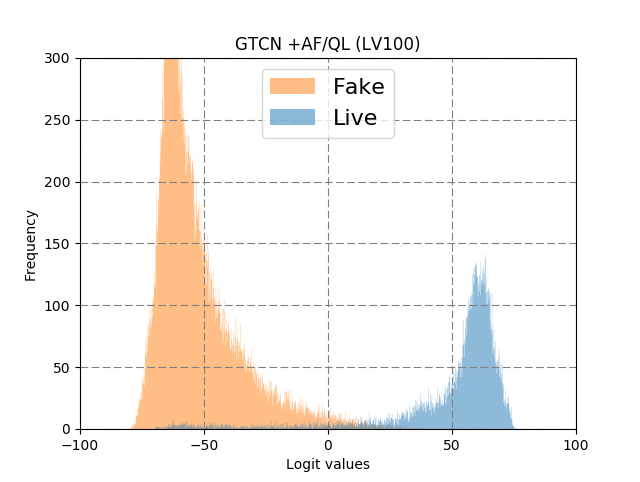}
    \hspace{-2mm}
    \includegraphics[width=6cm]{./supp/GTCN128_LV100_QLAFST}
    \hspace{-2mm}
}
\caption{Histograms of logit values for liveness test dataset. The compared models are trained on LV100 dataset. (1) GTCN Separate (2) GTCN +Joint (3) GTCN +AF (4) GTCN +QL (5) GTCN +AF/QL (6) GTCN +AF/QL/ST}
\label{fig:histogram_methods_liveness2}
\end{figure*}
\subsection{Characteristics for within and between classes}
Properly defining what we mean by visual similarity is both critical and complicated.
In quantitative terms, visual similarity of two classes can be measured by Fisher’s criterion score $J$~\cite{fisher1936use}.
Fisher's score $J$ in the experiment is defined as:
\begin{equation}\label{eq:eq1}
J = \frac{\left|\mu_{A}-\mu_{B}\right|^{2}}{\sigma_{A}^{2} + \sigma_{B}^{2}},
\end{equation}
where $\mu_{A}$ and $\mu_{B}$ are the mean of the logits for distribution A and B respectively. $\sigma_{A}$ and $\sigma_{B}$ denote the standard deviations of the same logits.
If the score is smaller than a specific threshold, we define the two classes as being similar and hard to discriminative.
In qualitative terms, we assume that visual similarity can be decomposed into two main aspects: style and underlying structure.
The comparison of Fisher's criterion scores for face liveness detection are shown in Figure ~\ref{fig:FISHER_SCORE_LV}.
The proposed GTCN shows the best score, while other generative models show better scores than baseline CNNs.
However, BC achieves lower score than the CNNs.
In the ablation study results, utilizing all the proposed methods, Joint/AF/QL/ST, results in the best score.
To clarify understanding for the scores, the histogram analysis of logit values for liveness test dataset are shown in Figure ~\ref{fig:histogram_methods_liveness1} and Figure ~\ref{fig:histogram_methods_liveness2}.
CNN256$\times$256 seems to outperform CNN128$\times$128.
In case of BC, intra-class variance and mean difference between classes are both reduced.
In VAE and GAN, intra-class variance and inter-class margin are both increased.
The proposed GTCN enlarges margin well beween live class and fake class.
In the ablation study, Joint, AF, QL, and ST show different characteristics to form the class distribution.
Joint learning makes smaller variance within each class than Separate learning, while AF enlarges margin between classes.
QL apparently reduces variance of each class, while it enlarges margin between the classes.
Additionally, we provide the results of a t-distributed stochastic neighbor embedding t-SNE~\cite{maaten2008visualizing} visualization in Figure ~\ref{fig:TSNE_TRAINING}.
Since GTCNs produce translated data from given real data, the feature space of augmented training samples are visualized.
The feature space of the classifiers for test samples of the face liveness dataset is also visualized.
The t-SNE of the GTCN with our proposed methods results in a sharper distinction between examples of similar classes, highlighting the fact that the learned representation is of higher quality.

\begin{figure}[!t] 
  \centering
  \centerline{\includegraphics[width=7.5cm]{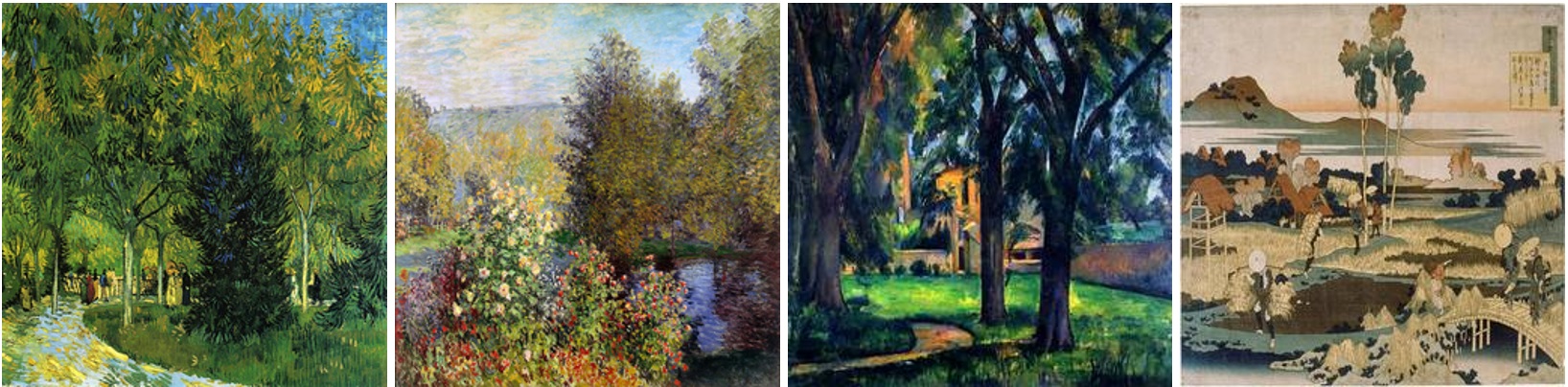}}
  \caption{Example from artist datasets.
           The first column shows Van Gogh's painting. The second, third, and fourth column show Monet, Cezanne, and Ukiyoe's one.}
  \label{fig:DATASET_EXAMPLE_MULTI}
\end{figure}

\begin{figure}[!t] 
    \centering
    \centerline{
        {\includegraphics[width=2.9cm]{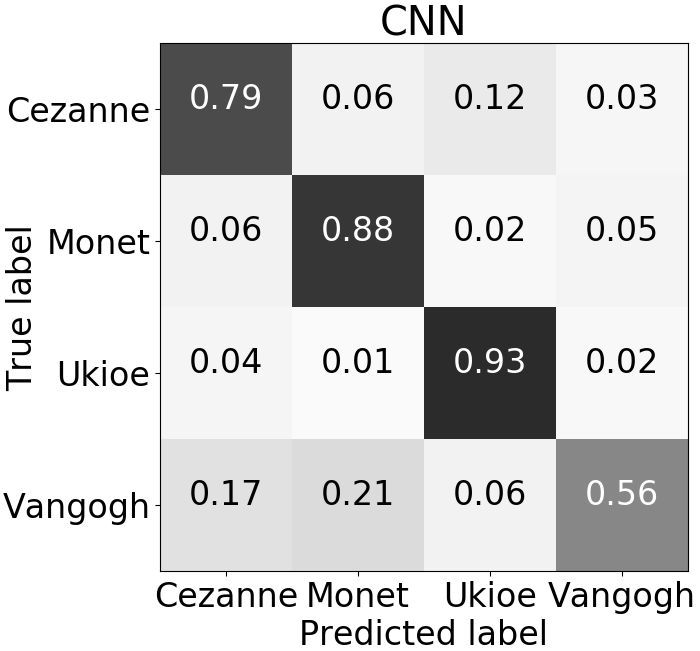} }%
        \hspace*{-0.2cm}
        {\includegraphics[width=2.9cm]{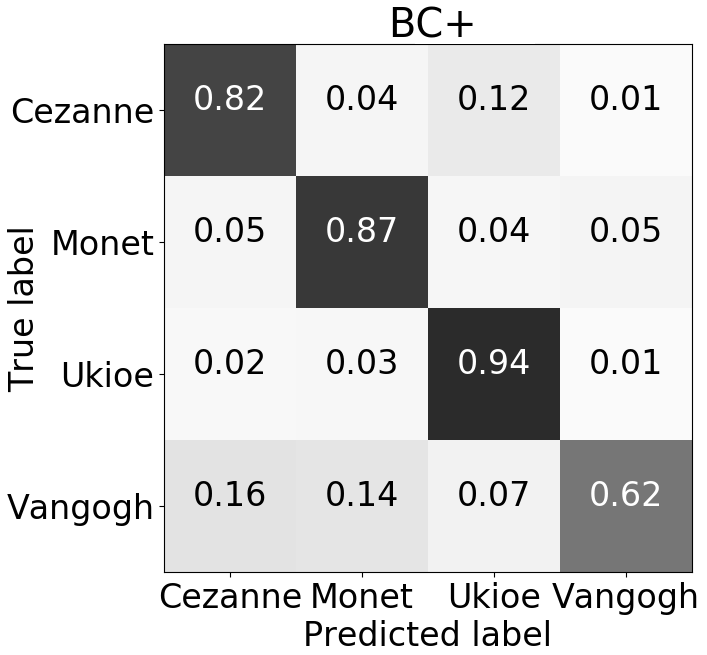} }%
        \hspace*{-0.2cm}
        {\includegraphics[width=2.9cm]{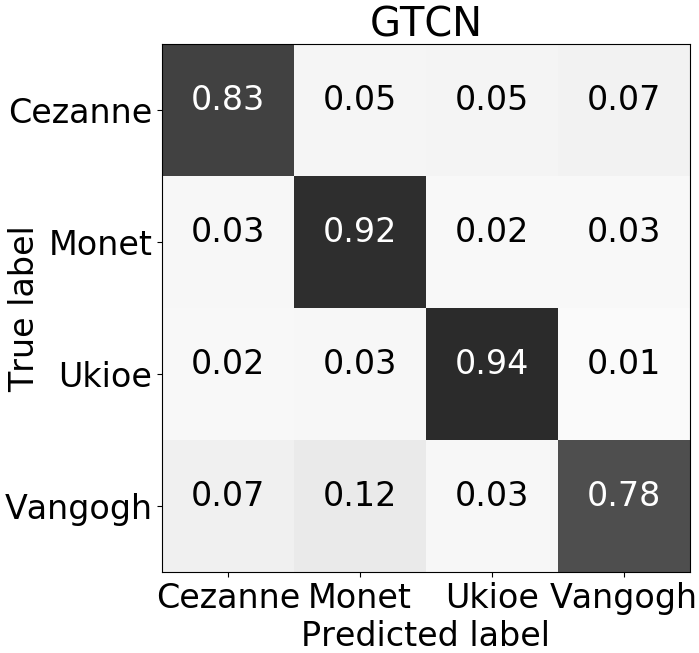} }%
    }
    \caption{Comparison of confusion matrix for compared methods on the artist dataset}
    \label{fig:ARTIST_CF}
\end{figure}

\begin{table}[!t] 
  \caption{Configuration of experimental datasets and subsampled variants.
           AT{*} are reconfigured training datasets.}
  \label{tab:db_define_multi}
  \centering
  \small
  \tabcolsep=0.18cm
  \scalebox{0.9}{
  \begin{tabular}{lccccc}
    \toprule
    Artist              & AT100    & AT80     & AT60     & AT40    & Test \\
    \midrule
    Cezanne             & 291      &  231     & 173      & 115     & 292  \\
    Monet               & 597      &  477     & 358      & 238     & 586  \\
    Ukiyoe              & 412      &  329     & 247      & 164     & 413  \\
    Van Gogh            & 200      &  160     & 120      & 80      & 200  \\
    \bottomrule
  \end{tabular}
  }
  \normalsize
\end{table}

\begin{table}[!t] 
  \caption{Comparison of mean accuracy on multi-class artist dataset}
  \label{tab:ARTS_db_eval_cnn}
  \centering
  \small   
  \tabcolsep=0.48cm
  \scalebox{0.9}{
  \begin{tabular}{lcccc}
    \toprule
    Model            & AT100  & AT80  & AT60  & AT40 \\
    \midrule
    CNN              & 83.21  & 80.95 & 77.68 & 69.42 \\
    CNN-256          & 87.28  & 86.54 & 83.21 & 72.42 \\
    \midrule
    BC               & 83.08  & 82.41 & 78.95 & 69.62 \\
    BC$^{+}$          & 84.74  & 82.74 & 79.55 & 68.95 \\
    Semi-sup.        & 86.81  & 84.68 & 80.55 & 71.75 \\
    \midrule
    GTCN             & \bfseries88.81  & \bfseries86.81 & \bfseries81.55 & \bfseries75.62 \\
    \bottomrule
  \end{tabular}
  }
  \normalsize
\end{table}

\begin{table}[!t] 
  \caption{Ablation experimental results of the proposed methods on the artist datasets}
  \label{tab:ablation_study_multi}
  \centering
  \small
  \tabcolsep=0.1cm
  \scalebox{0.9}{
  \begin{tabular}{lcccccc} 
       \toprule
      Artist       & CNN   & Separate & +Joint   & +AF   & +AF/QL   & +AF/QL/ST \\
      \midrule
      Accuracy     & 83.21 & 82.61 &85.14  & 85.88  & 86.21    & \bfseries88.81 \\
      \bottomrule
  \end{tabular}
  }
 \normalsize
\end{table}

\begin{figure*}[!t]
\centering
\centerline{
    {\includegraphics[width=6cm]{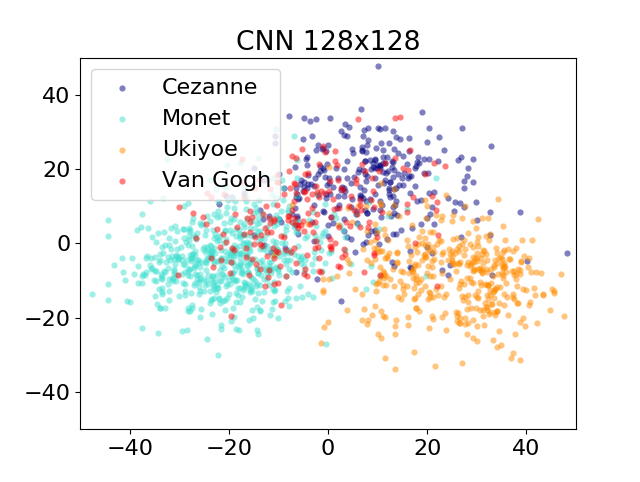}}
    \hspace{-2mm}
    {\includegraphics[width=6cm]{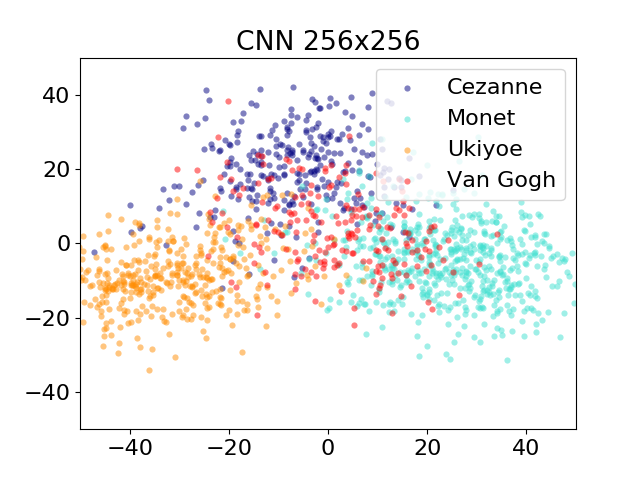}}
    \hspace{-2mm}
    {\includegraphics[width=6cm]{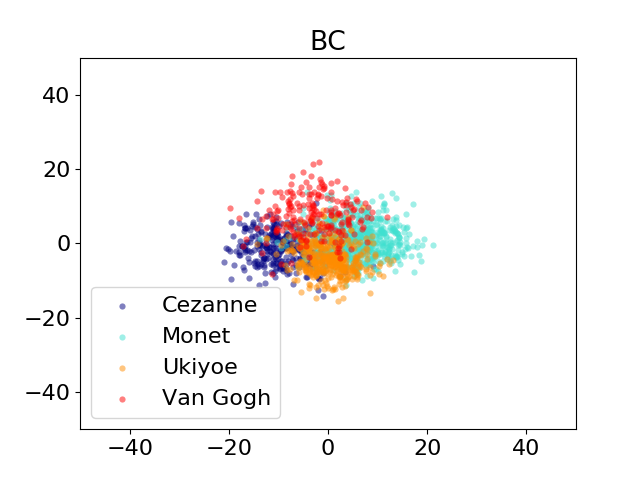}}
}
\centerline{
    {\includegraphics[width=6cm]{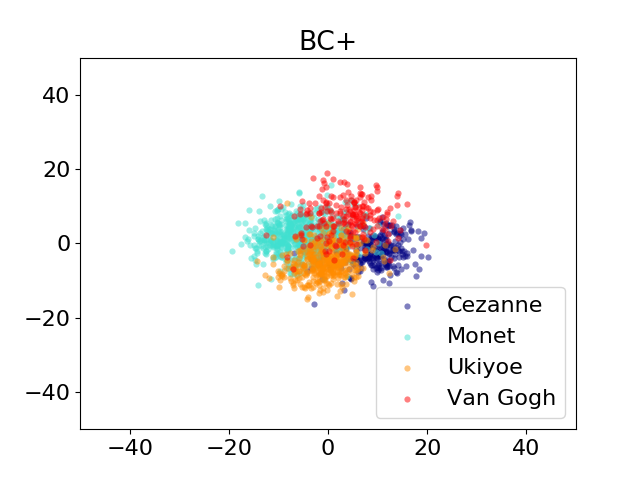}}
    \hspace{-2mm}
    {\includegraphics[width=6cm]{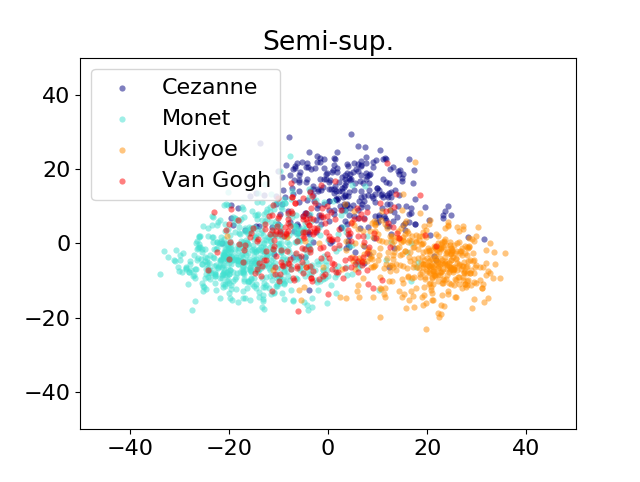}}
    \hspace{-2mm}
    {\includegraphics[width=6cm]{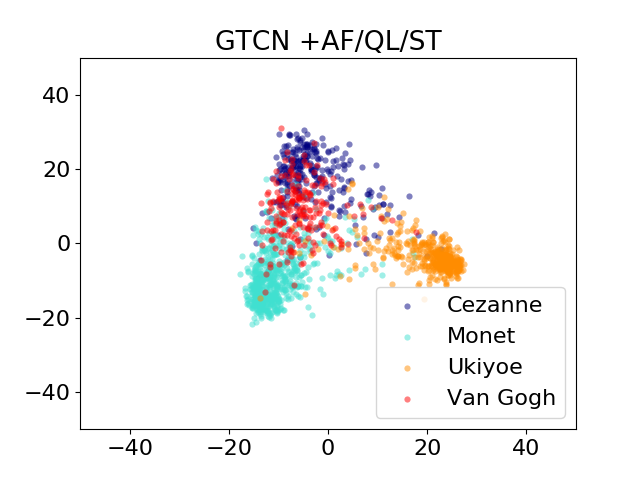}}
}
\caption{PCA analysis of logit values for artist test dataset. The compared models are trained on AT100 dataset. GTCN apparently enlarges margin between classes and reduce variance of within class. (1) CNN 128$\times$128 (2) CNN 256$\times$256 (3) BC 128$\times$128 (4) BC+ 128$\times$128 (5) Semi-sup. 128$\times$128 (6) GTCN 128$\times$128}
\label{fig:PCA_methods_artist1}
\centerline{
    {\includegraphics[width=6cm]{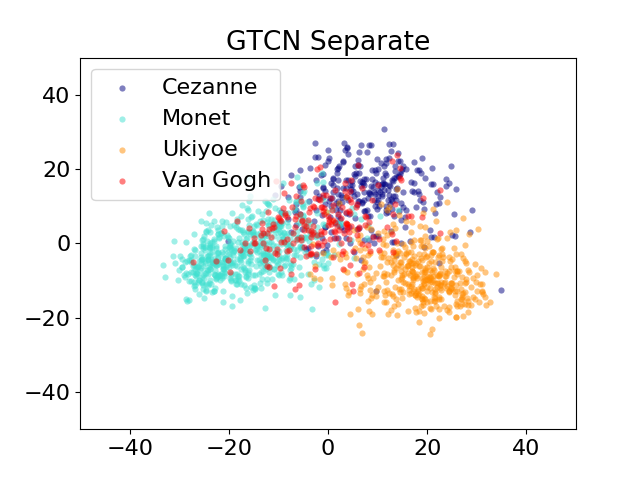}}
    \hspace{-2mm}
    {\includegraphics[width=6cm]{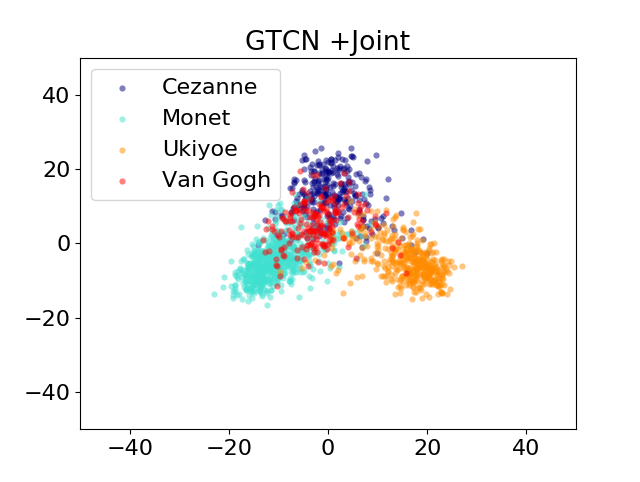}}
    \hspace{-2mm}
    {\includegraphics[width=6cm]{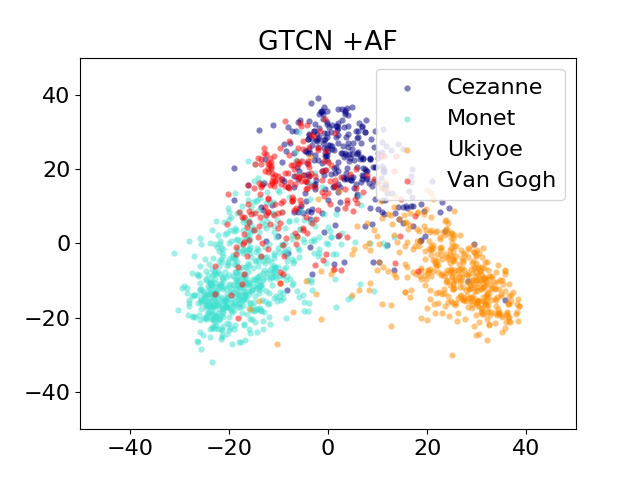}}
}
\centerline{
    {\includegraphics[width=6cm]{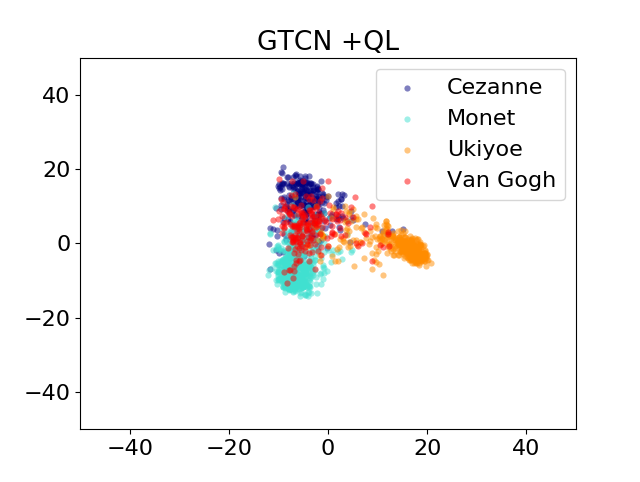}}
    \hspace{-2mm}
    {\includegraphics[width=6cm]{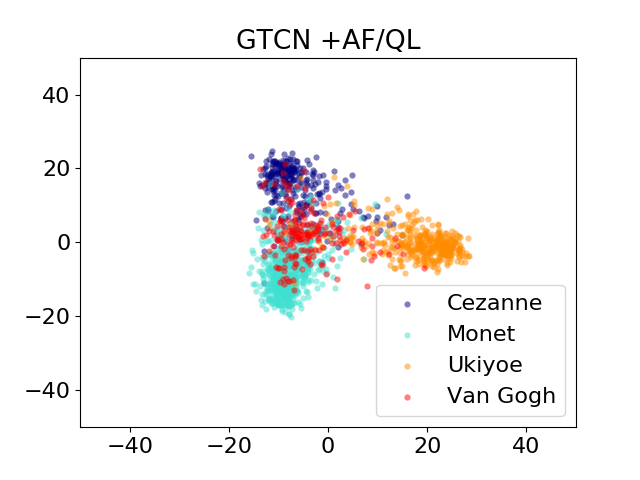}}
    \hspace{-2mm}
    {\includegraphics[width=6cm]{./supp/PCA_GTCN_AFQLST}}
}
\caption{PCA analysis of logit values for artist test dataset. The compared models are trained on AT100 dataset. GTCN +AF/QL/ST enlarges margin between classes and reduce variance of within class for transforming to separable distribution of four classes. (1) GTCN Separate (2) GTCN +Joint (3) GTCN +AF (4) GTCN +QL (5) GTCN +AF/QL (6) GTCN +AF/QL/ST}
\label{fig:PCA_methods_artist2}
\end{figure*}

\begin{figure}[!t] 
  \centering
  \centerline{\includegraphics[width=7.5cm]{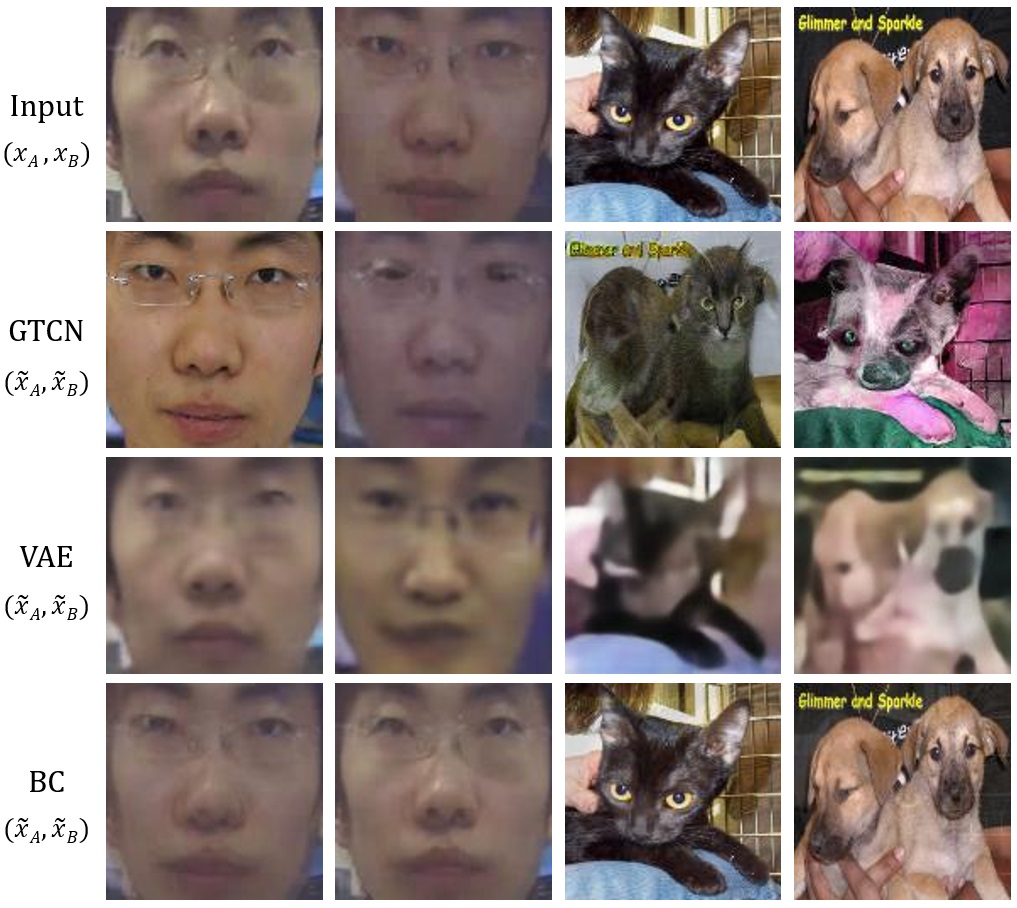}}
  \centerline{\includegraphics[width=7.5cm]{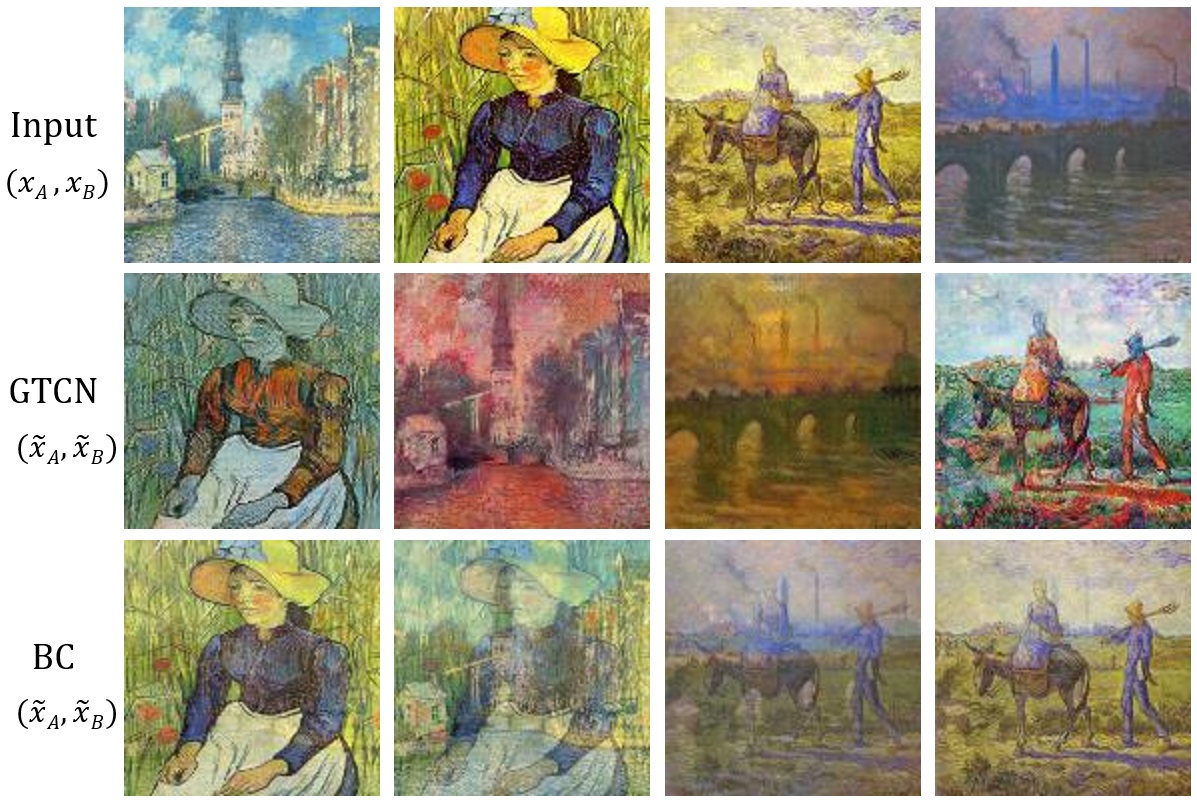}}
  \caption{Qualitative comparison of data augmentation methods to train a classifier.
   First four rows are binary class examples of the face liveness and the dogs vs. cats dataset.
   Last three rows are multi-class examples of the artist dataset.
   In the examples, BC uses 0.33 and 0.66 as mixing ratio between two images.}
  \label{fig:qualitative_images}
\end{figure}

\begin{figure}[!t] 
  \centering
  \centerline{\includegraphics[width=7.5cm]{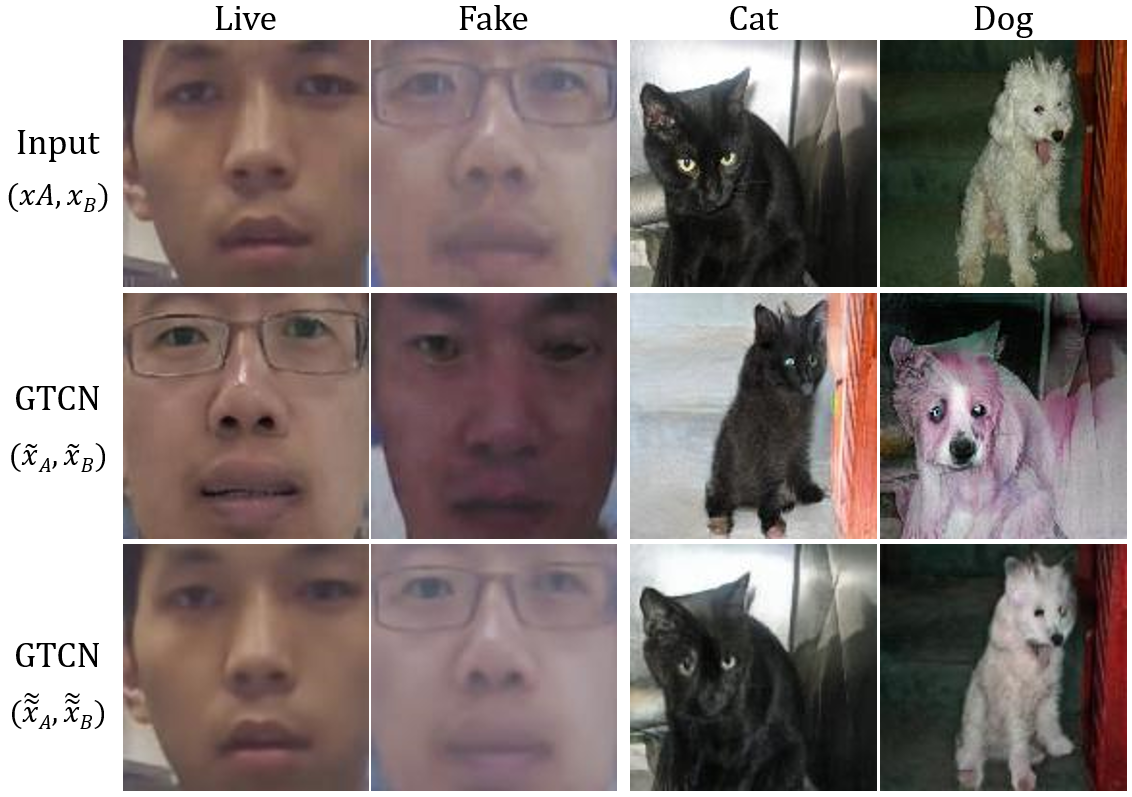}}
  \caption{Example input images, translated images, and cyclic reconstructed images of face liveness and dogs vs. cats datasets}
  \label{fig:qualitative2_images}
\end{figure}

\subsection{Multi-class style classification}
We extend the proposed method to a multi-class style classification setting.
As an example of the cases with lack of data, artists usually draw a limited number of paintings in their lifetime, which makes artist classification a challenging problem.
We constructed the artist dataset\footnote{\url{https://github.com/GTCN/datasets}\label{fn:dataset}}~that contains paintings from four painters as a multi-class classification experiment.
The specific configuration is described in Table ~\ref{tab:db_define_multi} and examples images of datasets are shown in Figure ~\ref{fig:DATASET_EXAMPLE_MULTI}.
Paintings have little overall structure, despite the styles chosen being similar, which creates a challenging problem of artist classification.
This requires a slight modification of our proposed approach.
$G_{AB}(\cdot)$ and $G_{BA}(\cdot)$ are trained as two translators across multiple-classes.
A mini-batch size is set to four and we sample examples from the mini-batch belonging to different classes.
Specifically, we use the same architecture as the one in the paper, which has only two generators, via a simple trick of setting minibach size to be four consisting of only one $x_A$, one $\tilde{x}_A$, one $x_B$, and one $\tilde{x}_B$ meaning that each minibatch handles only two classes at a time.
To calculate the probability for multi-class $k$ from a given test image $x$,
we employ the softmax output of the network.
The function is defined by
$P(y=i|x;\theta_{C})=  \frac{exp({FC_{i}})}{\sum_{j=1}^{k}exp({FC_{j}})}$,
where $1 \leq i \leq k$ is a class-id, $x$ is a given image, and $k$ is the number of classes.
We choose the class-id with the maximum probability among $k$ probabilities as the recognized class-id of $x$.
The hyperparameters for the quadruplet loss were set to: \{$\eta_{a}/\eta_{b}$=0.25, $\eta_{c}$=1.5\} for the artist dataset.
After training GTCNs, we additionally fine-tune the classifier with a cross-entropy loss for a few epochs.
The method quickly stabilizes and improves the overall accuracy of the classifier in GTCNs for a multi-class problem.
By fine-tuning, we assume weights related to noisy samples of multi-class may be truncated.
To perform a fair comparison, we tried applying fine-tuning to other compared methods such as CNNs and BC/BC$^{+}$s.
However, the fine-tuning is only useful for GTCN.
Evaluation results are shown in Table \ref{tab:ARTS_db_eval_cnn} and GTCNs outperform compared methods.
The accuracy for the four classes improved evenly for the GTCN as shown in Figure ~\ref{fig:ARTIST_CF}, because translated samples were used for training.
We perform an ablation study, the results of which are shown in Table \ref{tab:ablation_study_multi}.
Note that this is more like a simple modification for multi-class classification rather than carefully designed extension.
However, it is noteworthy that because of joint training of GAN and classifier, the proposed method achieves 88.81\% of accuracy for artist dataset significantly outperforming the baseline CNN (83.21\%) or separate training of GAN and CNN (82.61\%).
In terms of characteristics for within and between classes, 2D PCA of logit values for artist test dataset are shown in Figure ~\ref{fig:PCA_methods_artist1} and Figure ~\ref{fig:PCA_methods_artist2}, to analyze multi-class classification results.
Characteristics of comparison with other methods and ablation study are very similar to binary classification cases.
Especially, class Van Gogh has usually larger variance than other classes.
We suppose the reason of the overlapping with other artist classes is that he affected or was affected to other artist styles.
So, one of hardest classes is Van Gogh class in this problem.
Even though Vangogh is a hard class, GTCN shows apparently improvement to CNN and BC+ in Figure 12.
Overall results show the proposed GTCN enlarges margin between classes and reduce variance of within class.

\begin{figure*}[!t]
\centering
\centerline{{\includegraphics[width=17.5cm]{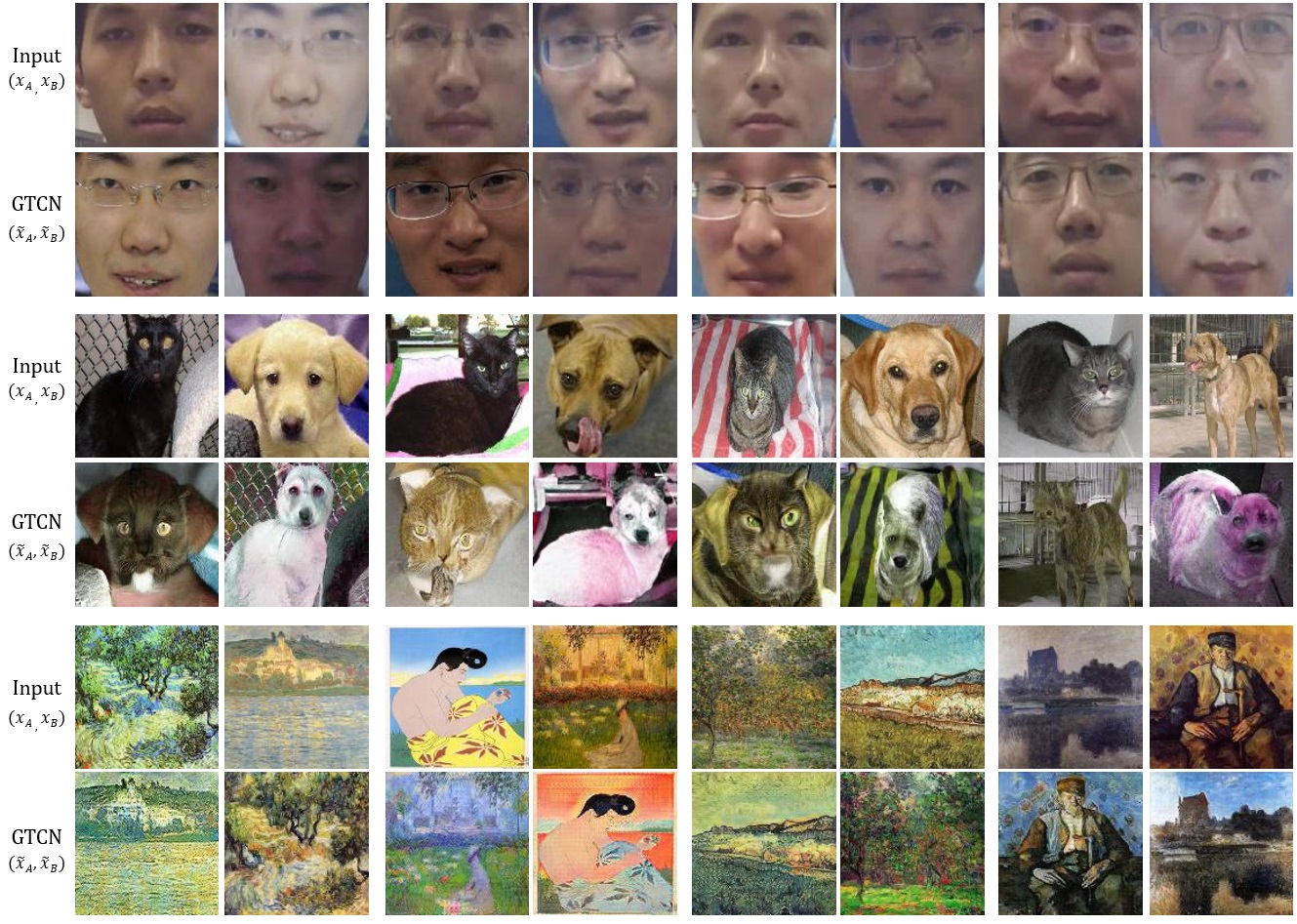}}}
\caption{Examples of augmented training images for GTCN: (Top) Face liveness dataset (Middle) Dogs vs. cats dataset (Bottom) Artist dataset
In the examples, translated samples attempt to borrow structure and shape information from samples of different classes while usually preserving texture and style.
Even though the translated images are sometimes visually weird, the images in the inter-class space also contribute to improving of the classifier's accuracy for visually similar images in terms of shape and style.
}
\label{fig:EXAMPLE_IMAGES}
\end{figure*}
\subsection{Visual examples of trained images}
The GTCN uses progressively translated images as a part of the mini-batches.
Note that the main purpose of the GTCN is not to generate realistic images but rather to improve the classifier's accuracy.
Nonetheless, a visual inspection of translated/generated samples gives insights into why certain methods work better than others.

As shown in Figure~\ref{fig:qualitative_images}, all of the compared models generate visually different images from given inputs to construct the augmented mini-batch.
The images generated by the VAE are blurry and seem to wrongly interpolate across classes.
BC attempts to mix up images, but seems to go pear-shaped at generating images that realistically belong to the target class, especially for the images of the artist dataset that display less overall structure.
However, the GTCN attempts to borrow structure and shape information from samples of different classes while usually preserve texture and style, so the classifier can learn with sharper and more diverse training data.
We visualize examples of input images, translated images, and cyclic reconstructed images for face and dogs vs. cats datasets in Figure~\ref{fig:qualitative2_images}.
To show more examples of augmented training images, visually informative pairs were chosen in Figure~\ref{fig:EXAMPLE_IMAGES}.

\section{Conclusion} \label{sec:Conclusion}
Our proposed model, GTCN, was trained with a novel joint learning approach.
In this work, a generative translation model and a classifier are trained via three innovations: the augmented mini-batch technique, adaptive fade-in learning, and the quadruple loss.
Since translators provide challenging data incrementally during training a classifier, accuracy can be improved by employing the augmented inter-class data.
After the end of training, we perform inference using only the light-weight classifier of GTCN.
Our method trained on a small subset of the whole dataset achieves a greater accuracy than the baselines trained on the full dataset.
When training on the full dataset, we surpass comparable state-of-the-art methods despite using low resolution images and a smaller network architecture for our method.
We believe our work can benefit classification tasks that suffer from visual similarity, diversity, and lack of data.


{
\bibliographystyle{ieee}
\bibliography{egbib}
}

\begin{IEEEbiography}[{\includegraphics[width=1in,height=1.25in,clip,keepaspectratio]{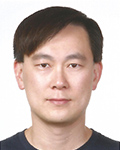}}]{ByungIn Yoo}
has been a Research Staff Member at Samsung Advanced Institute of Technology (SAIT), Suwon, Korea, since 2005.
He is currently a Principal Researcher and a Project Leader of Deep Image Processing Project in Computer Vision Lab. of SAIT.
Prior to joining Samsung, he conducted research projects on Distributed Systems, Military Simulators, and Computer Anti-virus Software from 1999 to 2005.
He received his B.S. and M.S. degrees in Computer Science and Engineering from Dongguk University, Seoul, Korea, in 1997 and 1999,
and the Ph.D. candidate at Statistical Inference and Information Theory lab in Korea Advanced Institute of Science and Technology (KAIST) under the supervision of Pr. Junmo Kim.
He was a Visiting Researcher at Montreal Institute for Learning Algorithms, Canada in 2018 on a scholarship from Samsung.
His research interests include computer vision and image processing based on exploiting pattern recognition and deep learning algorithms.
\end{IEEEbiography}

\begin{IEEEbiography}[{\includegraphics[width=1in,height=1.25in,clip,keepaspectratio]{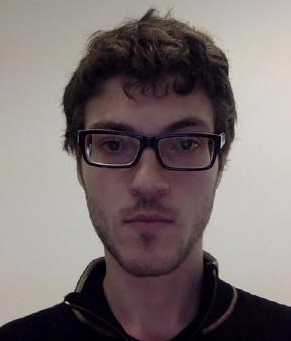}}]{Tristan Sylvain}
obtained his engineering diploma (equiv. MSc.) at Ecole Polytechnique in 2013, and MSc. in Computer Science at the University of Oxford in 2014. He is currently a Ph.D. student in Deep Learning at the Mila lab in Montréal under the supervision of Pr. Yoshua Bengio and Pr. Devon Hjelm. His research interests include Computer Vision, Generalization and Generative models. He currently focuses on understanding the fundamental principles that help generalization in different computer vision problems including zero-shot learning.
\end{IEEEbiography}

\begin{IEEEbiography}[{\includegraphics[width=1in,height=1.25in,clip,keepaspectratio]{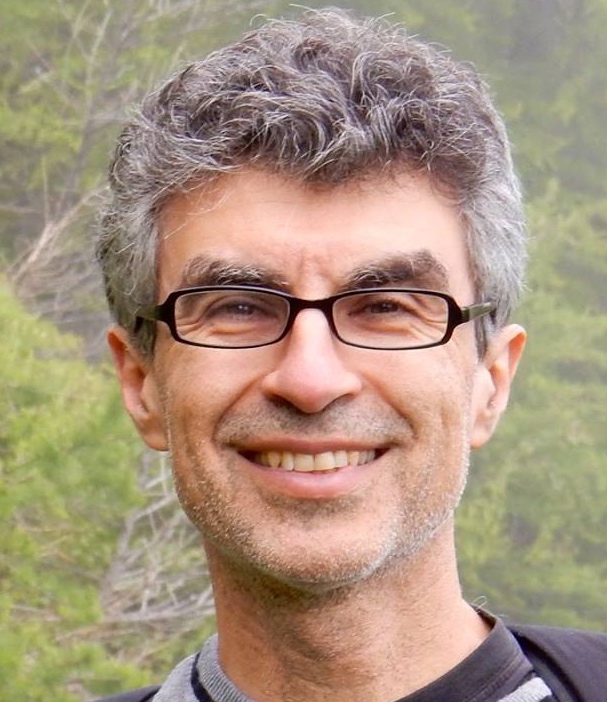}}]{Yoshua Bengio}
is recognized as one of the world’s leading experts in artificial intelligence and a pioneer in deep learning.
Since 1993, he has been a professor in the Department of Computer Science and Operational Research at the Université de Montréal. CIFAR’s Learning in Machines \& Brains Program Co-Director, he is also the founder and scientific director of Mila, the Quebec Artificial Intelligence Institute, the world’s largest university-based research group in deep learning.
In 2018, Yoshua Bengio ranked as the computer scientist with the most new citations worldwide, thanks to his many high-impact contributions.
In 2019, he received the Killam Prize and the ACM A.M. Turing Award, the Nobel Prize of Computing, jointly with Geoffrey Hinton and Yann LeCun for conceptual and engineering breakthroughs that have made deep neural networks a critical component of computing.
Concerned about the social impacts of this new technology, he actively contributed to the development of the Montreal Declaration for Responsible Development of Artificial Intelligence.
\end{IEEEbiography}

\begin{IEEEbiography}[{\includegraphics[width=1in,height=1.25in,clip,keepaspectratio]{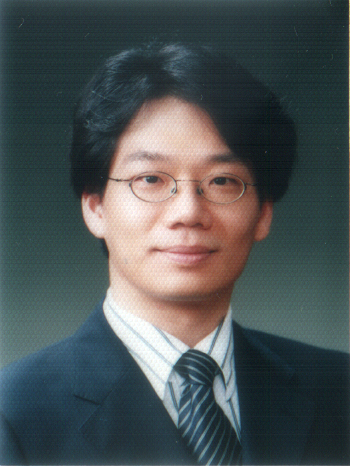}}]{Junmo Kim}
received the B.S. degree from Seoul National University, Seoul, South Korea, in 1998, and the M.S. and Ph.D. degrees from the Massachusetts Institute of Technology (MIT), Cambridge, MA, USA, in 2000 and 2005, respectively. From 2005 to 2009, he was with the Samsung Advanced Institute of Technology (SAIT), South Korea, as a Research Staff Member. He joined the faculty of KAIST, in 2009, where he is currently as an Associate Professor of electrical engineering. His research interests are in image processing, computer vision, statistical signal processing, machine learning, and information theory.
\end{IEEEbiography}
\EOD
\end{document}